\documentclass[acmsmall]{acmart}
\AtBeginDocument{%
  }

\acmISBN{978-1-4503-XXXX-X/2018/06}

\usepackage{algorithm}
\usepackage{algorithmicx}
\usepackage[noend]{algpseudocode}
\usepackage{diagbox}
\usepackage{subcaption}   
\usepackage{algpseudocode} 
\usepackage{xspace}
\usepackage{subcaption}

\usepackage{amsmath,mathtools,bm}

\usepackage[dvipsnames,svgnames,x11names]{xcolor}
\usepackage{graphicx}   
\usepackage{makecell}   
\usepackage{multirow}   
\usepackage{caption}  

\makeatletter
\newcommand{\removelatexerror}{\let\@latex@error\@gobble}
\makeatother

\theoremstyle{definition}

\theoremstyle{remark}

\theoremstyle{property}

\newcommand{\eat}[1]{}

\newcommand{\stitle}[1]{\vspace{.05in}\noindent{{\bf #1}}}

\newcommand{\rOneFC}[1]{\textcolor{black}{#1}}
\newcommand{\rTwoFC}[1]{\textcolor{black}{#1}}
\newcommand{\rThreeFC}[1]{\textcolor{black}{#1}}
\newcommand{\rAllFC}[1]{\textcolor{black}{#1}}

\newcommand{\name}{CoDec\xspace}

\renewcommand{\Comment}[1]{\textcolor{blue}{// \textit{#1}}}

\begin{document}
\setcopyright{cc}
\setcctype{by}
\acmJournal{PACMMOD}
\acmYear{2026} \acmVolume{4} \acmNumber{3 (SIGMOD)} \acmArticle{151}
\acmMonth{6} \acmDOI{10.1145/3802028}
\title{\name: Prefix-Shared Decoding Kernel for LLMs}
\author{Zhibin Wang}
\authornote{Both authors contributed equally to this research.}
\email{wzbwangzhibin@gmail.com}
\affiliation{%
  \institution{State Key Laboratory for Novel Software Technology, Nanjing University}
  \country{China}
}

\author{Rui Ning}
\authornotemark[1]
\email{rning@smail.nju.edu.cn}
\affiliation{%
  \institution{State Key Laboratory for Novel Software Technology, Nanjing University}
  \country{China}
}

\author{Chao Fang}
\email{chao.fang@kuleuven.be}
\affiliation{%
  \institution{State Key Laboratory for Novel Software Technology, Nanjing University}
  \country{China}
}

\author{Zhonghui Zhang}
\email{zhonghuizhang@smail.nju.edu.cn}
\affiliation{%
  \institution{State Key Laboratory for Novel Software Technology, Nanjing University}
  \country{China}
}

\author{Xi Lin}
\email{350904583lx@gmail.com}
\affiliation{%
  \institution{State Key Laboratory for Novel Software Technology, Nanjing University}
  \country{China}
}

\author{Shaobo Ma}
\email{shaoboma@smail.nju.edu.cn}
\affiliation{%
  \institution{State Key Laboratory for Novel Software Technology, Nanjing University}
  \country{China}
}

\author{Mo Zhou}
\email{221900277@smail.nju.edu.cn}
\affiliation{%
  \institution{State Key Laboratory for Novel Software Technology, Nanjing University}
  \country{China}
}

\author{Xue Li}
\email{youli.lx@alibaba-inc.com}
\affiliation{%
  \institution{Alibaba Group}
  \country{China}
}

\author{Zhongfeng Wang}
\email{fantasysee@foxmail.com}
\affiliation{%
  \institution{State Key Laboratory for Novel Software Technology, Nanjing University}
  \country{China}
}

\author{Chengying Huan}
\email{huanzhizun888@126.com}
\affiliation{%
  \institution{State Key Laboratory for Novel Software Technology, Nanjing University}
  \country{China}
}

\author{Rong Gu}
\authornote{Corresponding author.}
\email{gurong@nju.edu.cn}
\affiliation{%
  \institution{State Key Laboratory for Novel Software Technology, Nanjing University}
  \country{China}
}

\author{Kun Yang}
\email{kunyang@nju.edu.cn}
\affiliation{%
  \institution{State Key Laboratory for Novel Software Technology, Nanjing University}
  \country{China}
}

\author{Guihai Chen}
\email{gchen@nju.edu.cn}
\affiliation{%
  \institution{State Key Laboratory for Novel Software Technology, Nanjing University}
  \country{China}
}

\author{Sheng Zhong}
\email{sheng.zhong@gmail.com}
\affiliation{%
  \institution{State Key Laboratory for Novel Software Technology, Nanjing University}
  \country{China}
}

\author{Chen Tian}
\email{tianchen@nju.edu.cn}
\affiliation{%
  \institution{State Key Laboratory for Novel Software Technology, Nanjing University}
  \country{China}
}

\renewcommand{\shortauthors}{Zhibin Wang et al.}

\begin{abstract}
  Prefix-sharing among multiple prompts presents opportunities to combine the operations of the shared prefix, while attention computation in the decode stage, which becomes a critical bottleneck with increasing context lengths, is a memory-intensive process requiring heavy memory access on the key-value (KV) cache of the prefixes.
  Therefore, in this paper, we explore the potential of prefix-sharing in the attention computation of the decode stage.
  However, the tree structure of the prefix-sharing mechanism presents significant challenges for attention computation in efficiently processing shared KV cache access patterns while managing complex dependencies and balancing irregular workloads.
  To address the above challenges, we propose a dedicated attention kernel to \underline{co}mbine the memory access of shared prefixes in the \underline{dec}oding stage, namely \name.
  \name delivers two key innovations: a novel shared-prefix attention kernel that optimizes memory hierarchy and exploits both intra-block and inter-block parallelism, and a comprehensive workload balancing mechanism that efficiently estimates cost, divides tasks, and schedules execution.
  Experimental results show that \name achieves an average $1.9\times$ speedup and $120.9\times$ memory access reduction compared to the state-of-the-art FlashDecoding kernel regarding attention computation in the decode stage and $3.8\times$ end-to-end time per output token compared to the vLLM.
\end{abstract}

\ccsdesc[500]{Computing methodologies~Machine learning}
\ccsdesc[500]{Computing methodologies~Parallel algorithms}
\ccsdesc[500]{Information systems~Data management systems}


\keywords{GPU kernel, large language models (LLMs), attention computation, prefix-sharing}

\received{October 2025}
\received[revised]{January 2026}
\received[accepted]{February 2026}

\maketitle

\noindent\textbf{Code Availability:} Code is available at https://github.com/wzbxpy/codec.

\noindent\textbf{Earlier Version:} The earlier arXiv version of this paper was titled Flashforge.
\section{Introduction}
\label{sec:intro}

Large language models (LLMs) have demonstrated significant performance across diverse tasks, such as question answering~\cite{kamalloo-etal-2023-evaluating}, planning~\cite{significantgravitas2023autogpt,song2023llmplannerfewshotgroundedplanning}, code generation~\cite{github2024copilot, roziere2023codellama}, recommendation systems~\cite{Hou2023LargeLM,zhao2024recommendersystemseralarge}, and even solving complex mathematical problems~\cite{openai2024reasonwithllm, qwen2024qwq, deepseekai2024deepseekr1list, wu2024inferencescalinglawsempirical}. Despite their impressive capabilities, inference efficiency remains critical for LLM deployment, as it directly impacts user experience and operational costs~\cite{zhou2024surveyefficientinferencelarge,fang2025anda}. For instance, real-time applications such as chatbots and interactive AI agents require low-latency responses to maintain seamless interactions~\cite{luo2025largelanguagemodelagent}. Meanwhile, enterprise-scale deployments, where LLMs process millions of queries daily, must optimize computational resources to remain cost-effective~\cite{vllm_sosp23,flashdecoding,ma2025apt}.

\begin{figure}[tb]
    \centering
    \begin{subfigure}{0.35\linewidth}
        \centering
        \includegraphics[width=1.0\linewidth]{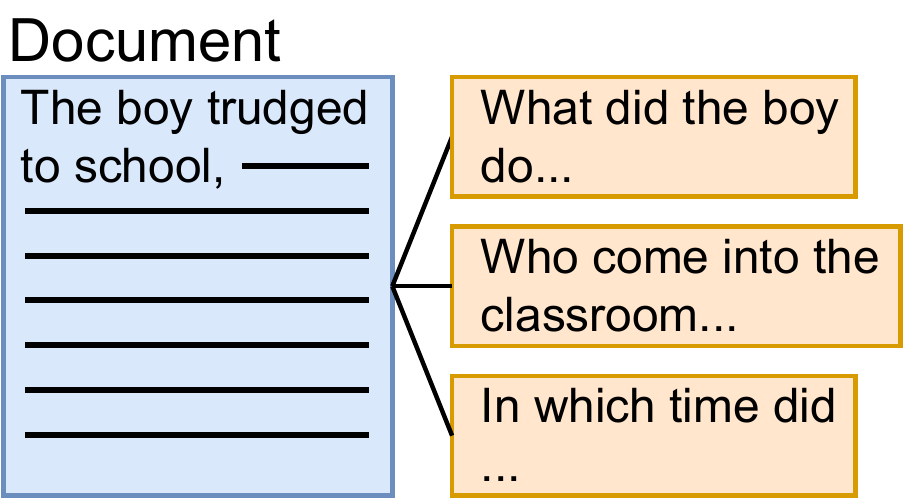}
        \caption{Prefix-shared document QA.}
    \end{subfigure}
    \hspace{0.05\linewidth}
    \begin{subfigure}{0.35\linewidth}
        \centering
        \includegraphics[width=1.0\linewidth]{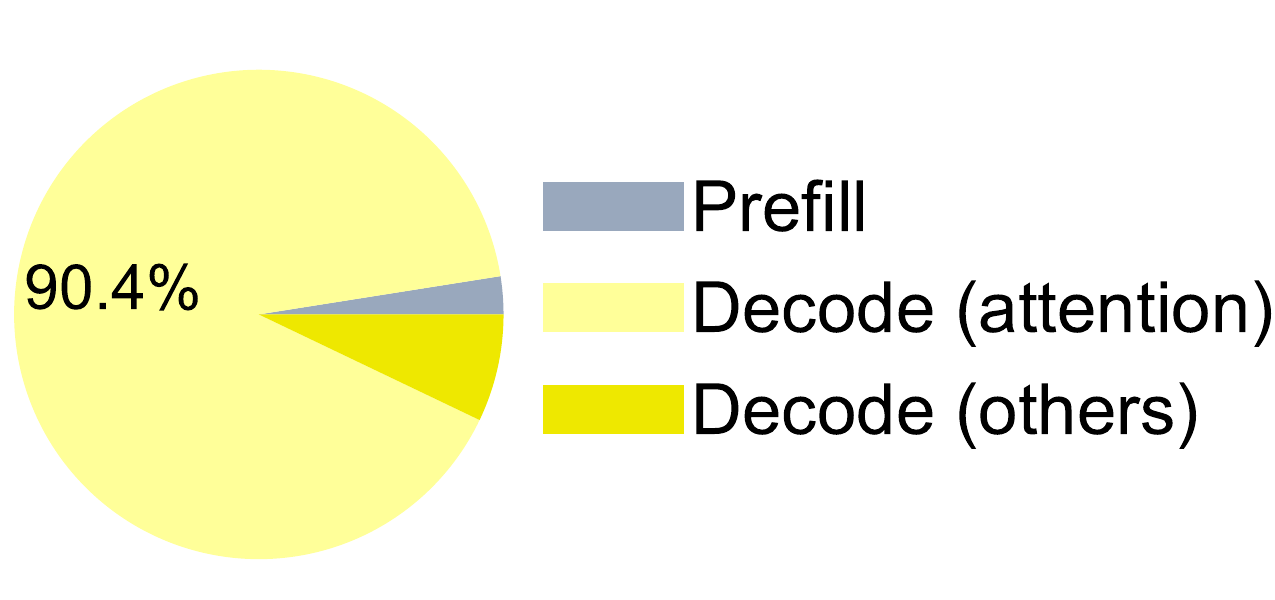}
        \caption{Cost of various operations.}
    \end{subfigure}
    \caption{Motivation of leveraging prefix-sharing in attention of decoding stage.}
    \label{fig:intropic}
    \vspace{-2em}
\end{figure}
One of the most promising directions for improving LLM inference efficiency is prefix-sharing founded on the observation that many prompts share identical prefixes, which is common in document question answering~\cite{eccleston2023sharegpt,li2024looglelongcontextlanguagemodels,xiao2021nextqanextphasequestionansweringexplaining}, tree-of-thoughts~\cite{yao2023treethoughtsdeliberateproblem}, speculative decoding~\cite{specinfer}, and few-shot prompting~\cite{reynolds2021prompt}.
For example, as shown in Figure~\ref{fig:intropic} (a), in document-based question answering, multiple questions may pertain to the same document ~\cite{eccleston2023sharegpt,li2024looglelongcontextlanguagemodels,xiao2021nextqanextphasequestionansweringexplaining}.
For prefill stage generating the KV cache for the prompt tokens, prefix-sharing can be utilized to reduce the duplicated computation and memory consumption of the KV cache generation of the shared prefix across different requests~\cite{zheng2024sglangefficientexecutionstructured, vllm_apc, zheng2024batchllmoptimizinglargebatched,zhao2024blendserveoptimizingofflineinference,srivatsa2024prebleefficientdistributedprompt,qin2024mooncakekvcachecentricdisaggregatedarchitecture,wang2025echoefficientcoschedulinghybrid}.

With increasing context lengths, the decoding stage, particularly attention computation, becomes the critical bottleneck in LLM inference, accounting for 90\% of total time as shown in Figure~\ref{fig:intropic} (b) when running 100K prompts with 128 output tokens on Llama-3.1-8B~\cite{huggingfaceMetallamaLlama8B}.
Different from the prefill stage generating the KV cache, the decode stage autoregressively generates the output tokens based on the generated KV cache, thereby sequentially processing the tokens one by one and requiring heavy memory access to the KV cache, exhibiting insufficient parallelism and memory-bound pattern~\cite{fu2024break}.
Recently, state-of-the-art attention optimization methods, such as FlashAttention~\cite{dao2023flashattention2} and FlashDecoding~\cite{flashdecoding}, have achieved significant speedups by leveraging shared memory (i.e., on-chip memory) to reduce memory access overhead and by increasing parallelism in attention computation.
Consider a batch of multi-head attention that takes three 4D tensors as input: the query ($\mathbf{Q}\in \mathbb{R}^{bs\times n_q \times h \times d}$), key ($\mathbf{K} \in \mathbb{R}^{bs\times n \times h \times d}$), and value ($\mathbf{V} \in \mathbb{R}^{bs\times n \times h \times d}$) tensors\footnote{The notation of the tensors is shown in Table~\ref{tab:notation}.}.
FlashAttention and FlashDecoding decompose the attention operation in the \emph{batch, head, query sequence, and KV sequence dimensions} into multiple blocks so that each block fits in shared memory and exposes four-way parallelism.
However, when employing these techniques in prefix-sharing scenarios, queries sharing identical prefixes are processed individually by separate computational units, even though they could potentially share memory access for the common KV cache.
This processing pattern inevitably results in duplicated memory transactions for accessing the shared KV cache.

\rAllFC{From a data management perspective, the KV cache is a large, dynamic, and performance-critical data structure: serving systems must decide how to lay out, reuse, and access KV blocks efficiently under strict latency constraints. CoDec targets the decode stage where KV access becomes bandwidth-bound, and treats shared-prefix attention as a data-access optimization problem---reducing redundant reads to shared KV blocks by coalescing access across requests. This is complementary to higher-level KV cache management and scheduling work (e.g., HotPrefix~\cite{hotprefix} and Cache-Craft~\cite{cachecraft}), which focuses on what to keep and when to reuse; CoDec focuses on how to execute the shared-prefix accesses efficiently once such reuse opportunities exist.}

In this paper, we explore leveraging prefix-sharing in the decode stage, specifically by optimizing the memory access patterns for shared KV cache across different requests.
This approach directly addresses the redundant memory transactions identified above, targeting the primary performance bottleneck in LLM inference.
However, when shifting the regular attention computation~\cite{flashdecoding,dao2023flashattention2} between 4D tensors to the irregular shared prefix attention computation, there are several challenges to be addressed:

\textbf{Challenge 1: Organizing complex dependencies in shared prefix attention computation.}
The management of the shared prefix KV cache has been well studied in the prefill stage~\cite{zheng2024sglangefficientexecutionstructured,yao2025cacheblendfastlargelanguage}, logically organized as a radix tree of 3D tensors where each node represents a chunk of the prefix KV cache.
However, further extending this tree structure to attention computation introduces two major issues: First, it requires not only the KV cache but also the corresponding query tensor shared with the prefix to coordinate KV cache access, complicating management. Second, reduction operations are needed to combine decoding results in corresponding prefix KV cache nodes for each query in the tree structure, which is nontrivial to be parallelized.
Prior art~\cite{ye2024chunkattentionefficientselfattentionprefixaware, cascade-inference} only considered the trivial case where all requests share the same prefix, thus simply handling shared and non-shared computations separately.
Hence, efficiently leveraging the KV cache tree structure for both attention and reduction operations remains challenging yet essential.

\textbf{Challenge 2: Balancing the workload of irregular prefix-shared attention computation.}
The workload for attention computation between each KV cache node and its corresponding query tensor is determined by both query count and prefix length of the KV cache node, which vary significantly across computations, leading to highly irregular workloads~\cite{oh2024exegpt}.
Moreover, varying shapes of query tensors and prefix KV cache nodes result in divergent memory or compute bounds~\cite{10.1145/3676641.3716243, 10.1145/1498765.1498785},  making theoretical workload estimation impractical.
This necessitates an integrated solution combining cost estimation, intelligent task division, and efficient scheduling to balance workloads without resorting to expensive fine-grained partitioning.

To address the above challenges, we develop the dedicated prefix-shared attention operator, namely \name, which \underline{co}mbines the memory access of the attention computation of the shared prefix across different requests in the \underline{dec}ode stage.
\name delivers two key innovations: First, it implements a novel shared-prefix attention kernel that optimizes both memory hierarchy between shared and global memory and exploits intra-block and inter-block parallelism.
Second, it incorporates a comprehensive workload balancing mechanism with a cost estimator, task divider, and scheduler that guides execution before the attention kernel runs.
\rTwoFC{It should be noted that \name follows the same paged KV-cache layout as PagedAttention and exposes an attention interface compatible with FlashDecoding, enabling straightforward integration into existing serving stacks such as vLLM.}
We summarize the contributions of this paper as follows:

\stitle{Shared prefix attention kernel (Section~\ref{sec:kernel}).}
To efficiently form the attention computation, we introduce the indexes between the prefix KV cache tree and the query tensors, which facilitate loading the corresponding tensors to the shared memory. Moreover, we abstract two fundamental primitives in the block-level, i.e., \emph{partial attention computation (PAC)} to compute the partial output between 2D query and KV tensors extracted from global query and KV tensors, and \emph{partial output reduction (POR)} to reduce two partial outputs of the same query. Building on these two primitives, we propose an inter-block computation task executor and a dedicated tree-based reduction. The tree-based reduction aims to maximize the GPU utilization by achieving a parallelism degree equal to the block number while minimizing the number of reduction operations.

\stitle{Workload balancing mechanism (Section~\ref{sec:wb}).}
Directly mapping the partial attention computation between each KV cache node and the corresponding query tensor suffers from significant irregular workload and insufficient parallelism. Therefore, we further divide the computation into multiple subtasks. Recognizing that the workload of each subtask is neither determined by IO complexity nor compute complexity as shown in Table~\ref{tab:block_execution_time}, we propose a profile-based cost estimator to guide the task division. Moreover, we formulate the optimization problem of task division and scheduling, unfortunately, it is NP-hard. However, given the specific characteristics of partial attention computation, i.e., coarse-grained has less overhead than fine-grained, we propose a greedy algorithm to solve the problem.

\stitle{Evaluation (Section~\ref{sec:eval}).}
\rAllFC{We evaluate \name across a range of prefix-sharing workloads, GPUs, attention variants, and model sizes.} Compared to FlashDecoding, \name achieves an average $1.9\times$ speedup and a $120.9\times$ reduction in global memory access for decode-stage attention, and achieves a $3.8\times$ TPOT speedup over vLLM in end-to-end comparison.

\section{Background} \label{sec:background}
\subsection{LLM Inference}

\begin{figure}[htb]
    \centering
    \includegraphics[width=1\linewidth]{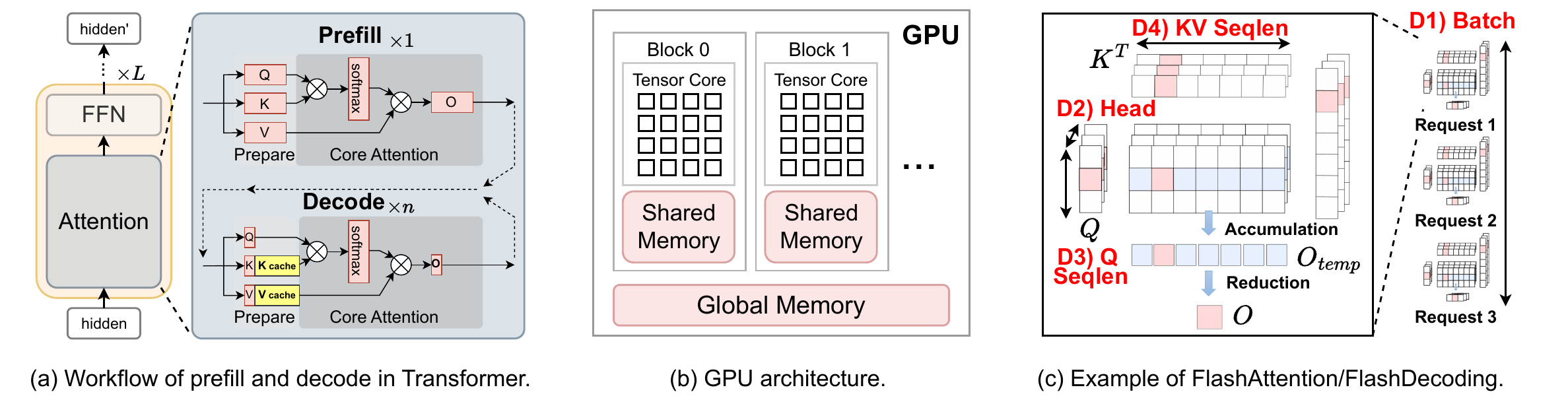}
    \caption{Background knowledge of LLM inference, GPU architecture, and existing attention kernels. }
    \label{fig:background}
\end{figure}

\stitle{Transformer architecture:}
Recent mainstream LLMs, such as ChatGPT~\cite{openai2024chatgpt},  DeepSeek~\cite{deepseekai2025deepseekv3technicalreport}, Llama~\cite{grattafiori2024llama3herdmodels}, and Gemini~\cite{geminiteam2024geminifamilyhighlycapable} are based on the transformer architecture~\cite{attentionisallyouneed}, which generates tokens in an auto-regressive manner.
As shown in Figure~\ref{fig:background} (a), the transformer consists of the attention module and the feed-forward network (FFN) module. The attention module calculates the attention scores between each pair of tokens, allowing the model to learn the relationships and dependencies among them, which makes the transformer model outperform the RNN model~\cite{lipton2015criticalreviewrecurrentneural} considering the long-range dependencies~\cite{attentionisallyouneed}.
The FFN module is responsible for learning complex representations of the tokens.
These two modules are typically stacked $L$ times to deepen the model.

\stitle{Prefill and decode stages:}
The inference process of these models comprises two primary stages: prefill and decode, which are depicted in Figure~\ref{fig:background} (a). During the prefill stage, the model simultaneously processes the input token sequence, caches the corresponding key and value (KV) tensors for these tokens, and generates the initial output token. In the subsequent decode stage, the model processes one token per step, generating the next token autoregressively based on previously generated tokens and the cached KV tensors. Because the prefill stage processes numerous tokens concurrently, it is computationally intensive and typically compute-bound. Conversely, the decode stage processes only a single token per step, resulting in significantly lower computational demands and rendering this stage memory-bound.

Figure~\ref{fig:intropic} (b) presents the prefill/decode time breakdown of two workloads when serving Llama-3.1-8B. When running 100K prompts with 128 output tokens, the overall decoding latency is 102s, while the prefill latency is only 2.62s, and the attention kernel accounts for 90\% of the total time.
Furthermore, as sequence length increases, the proportion of time consumed by the attention kernel also rises.

In summary, with the increasing context length, \emph{the attention computation in the decode stage becomes the critical bottleneck of the LLM inference process}. Therefore, it is crucial to optimize the attention computation in the decode stage.

\begin{table}[htb]
    \centering
    \caption{Notation Table.}
    \begin{tabular}{c|p{0.8\linewidth}}
        \toprule
        \textbf{Notation} & \textbf{Definition}                               \\
        \hline
        $Q,K,V,O$         & Partial 2D Query, key, value and output tensor    \\
        $\textbf{Q},\textbf{K},\textbf{V},\textbf{O}$
                          & Query, key, value and output tensor of batch      \\
        $S,P$             & Attention score before and after softmax function \\
        $h, d$            & Number of attention heads, hidden layer dimension \\
        $bs,n,n_q$        & Batch size, sequence length, query tokens' length \\
        $T$               & Set of tasks                                      \\
        $p,t$             & Number of parallel thread blocks and tasks        \\
        $b_q,b_k$         & Number of slices in the query, KV cache dimension \\
        \bottomrule
    \end{tabular}
    \vspace{-1em}
    \label{tab:notation}
\end{table}
\subsection{Attention Mechanism}
\label{sec:attention background}
Given the significance of the attention mechanism in both algorithm-level and system-level, we briefly review the self-attention operation and its variants in execution.

\stitle{Self-attention operation:}
The key idea of the self-attention operator is to compute the attention score between each pair of tokens $i$ and $j$, which indicates the importance of token $j$ in the context of token $i$. Subsequently, the embedding of token $i$ is updated by a weighted sum of the embeddings of all tokens, where the weights are determined by the attention scores.

Formally, given a sequence of input tokens, three linear transformations generate the query ($Q\in \mathbb{R}^{n_q \times d}$), key ($K \in \mathbb{R}^{n \times d}$), and value ($V \in \mathbb{R}^{n \times d}$) tensors, where $n$ is the sequence length, $d$ is the hidden size, and $n_q$ is the sequence length of the query tokens. Note that in some cases $n_q\ne n$, such as in the decode stage or when leveraging chunked-prefill~\cite{agrawal2023sarathiefficientllminference}.
\begin{equation}
    \begin{aligned}
        O = \text{softmax}\left(\frac{QK^T}{\sqrt{d}}\right)V,
    \end{aligned}
\end{equation}
where $O\in \mathbb{R}^{n_q \times d}$ is the output tensor indicating the updated embedding of the query tokens. The softmax function is employed to normalize the attention scores of each query token, i.e., on each row. Given a vector $\mathbf{x}\in \mathbb{R}^n$, the softmax function is defined as:
\begin{equation}
    \begin{aligned}
        m=\max(\mathbf{x}),
        s=\sum_{i=1}^{n} e^{\mathbf{x}[i]-m},
        \text{softmax}(\mathbf{x})[i]=\frac{e^{\mathbf{x}[i]-m}}{s}.
    \end{aligned}
\end{equation}
$m$ is the maximum value of the vector $\mathbf{x}$ to prevent overflow in the exponentiation operation~\cite{dao2023flashattention2}.

In this paper, we use multi-head attention~\cite{attentionisallyouneed} as the default attention mechanism, and other attention mechanisms~\cite{shazeer2019fasttransformerdecodingwritehead,ainslie2023gqatraininggeneralizedmultiquery,deepseekai2024deepseekv2strongeconomicalefficient} will be discussed in Section~\ref{sec:related_work}.
Another consideration is batching, which is a common technique used to improve resource utilization by enlarging the workload of processing multiple requests in parallel.

Overall, the general attention operation for batched multi-head attention takes three input tensors: the query ($\mathbf{Q}\in \mathbb{R}^{bs\times n_q \times h \times d}$), key ($\mathbf{K} \in \mathbb{R}^{bs\times n \times h \times d}$), and value ($\mathbf{V} \in \mathbb{R}^{bs\times n \times h \times d}$) tensors.

\stitle{Attention variants in training, prefill, and decode stages:}
\begin{itemize}
    \item \textbf{Training} directly uses the general attention operation and $n_q=n$.
    \item \textbf{Prefill stage} usually sets the batch size $bs=1$ as there is enough workload in the prefill stage.
    \item \textbf{Decode stage} only processes one token for each request at a time, i.e., $n_q=1$, which \emph{suffers from both insufficient parallelism and lower arithmetic intensity (operations per memory access)~\cite{10.1145/3676641.3716243, 10.1145/1498765.1498785} and thus is memory-bound}.
\end{itemize}

\subsection{GPU Architecture}
\label{sec:gpu architecture}
We abstract the GPU architecture as shown in Figure~\ref{fig:background}(b), which logically consists of multiple blocks, each containing a tensor core, and a fast but limited on-chip memory (shared memory), while the large but slow global memory is shared among all blocks~\cite{nvidiaNVIDIAA100}. Therefore, two levels of parallelism can be exploited: 1) \textbf{intra-block parallelism} within each block, which is typically achieved through tensor cores and can be further accelerated by shared memory, and 2) \textbf{inter-block parallelism} across multiple blocks, which is needed to balance the workload among blocks.

\subsection{FlashAttention and FlashDecoding}
\label{sec:flashattention}

\stitle{FlashAttention:}
FlashAttention~\cite{dao2023flashattention2} and its successor FlashDecoding~\cite{flashdecoding} are highly optimized CUDA kernels for attention computation, designed to leverage the GPU's memory hierarchy and parallelism capabilities. As shown in Figure~\ref{fig:background}(c) and Algorithm~\ref{alg:flashattention} (Line~3-6), FlashAttention decomposes and parallelizes the attention operation in the \emph{1. batch, 2. head, 3. query sequence and 4. KV sequence dimensions} into multiple blocks, where each block can fit into the shared memory. Subsequently, in lines~7-11, the partial attention computation of each block is performed in shared memory, which takes $Q\in \mathbb{R}^{\frac{n_q}{b_q} \times d}$ and $K, V \in \mathbb{R}^{\frac{n}{b_k} \times d}$ to generate the partial output $O$. Finally, in line~12, we reduce the partial outputs in the KV sequence dimension to obtain the final output $O$.

\begin{algorithm}[t]
    \caption{FlashAttention}
    \label{alg:flashattention}
    \begin{algorithmic}[1]
        \State \textbf{Input:} $\mathbf{Q}, \mathbf{K}, \mathbf{V}$
        \State \textbf{Output:} $\mathbf{O}$
        \For{$seq = 1$ \textbf{to} $bs$ \textbf{in parallel}}
        \For{$head = 1$ \textbf{to} $h$ \textbf{in parallel}}
        \For{$i = 1$ \textbf{to} $b_q$ \textbf{in parallel}}
        \For{$j = 1$ \textbf{to} $b_k$ \textbf{in parallel}}
        \State \Comment{Executed in shared memory}
        \State $Q=\mathbf{Q}[seq, (i-1)\frac{n_q}{b_q}:i\frac{n_q}{b_q}, head, :]$
        \State $K=\mathbf{K}[seq, (j-1)\frac{n}{b_k}:j\frac{n}{b_k}, head, :]$
        \State $V=\mathbf{V}[seq, (j-1)\frac{n}{b_k}:j\frac{n}{b_k}, head, :]$
        \State $O[j] = \text{softmax}(\frac{QK^T}{\sqrt{d}})V$
        \EndFor
        \State $\mathbf{O}[seq, (i-1)\frac{n_q}{b_q}:i\frac{n_q}{b_q}, head, :] = \text{reduce}(O)$
        \EndFor
        \EndFor
        \EndFor
        \State \textbf{return} $O$
    \end{algorithmic}
\end{algorithm}

\emph{It is worth noting that FlashAttention and FlashDecoding are designed for regular 4D tensors, which makes parallelism and partitioning easier.}

\subsection{Prefix-sharing}
\label{sec:prefix-sharing}
Many real-world workloads exhibit opportunities for prefix sharing. Several notable examples include:
\begin{itemize}
    \item \textbf{Document Question Answering (QA)~\cite{eccleston2023sharegpt,li2024looglelongcontextlanguagemodels,xiao2021nextqanextphasequestionansweringexplaining}.} Users may ask multiple questions about the same document. For instance, in the LooGLE~\cite{li2024looglelongcontextlanguagemodels} dataset, the average prompt length is 23474 tokens, and the sharing rate is 91\%.
    \item \textbf{Tool-use~\cite{guo2024stabletoolbenchstablelargescalebenchmarking}.} If multiple requests share the same tool usage, they can share the prefix of the tool description and the tool usage instructions. The ToolBench dataset~\cite{guo2024stabletoolbenchstablelargescalebenchmarking} has an average prompt length of 1835 tokens and a sharing rate of 85\%.
    \item \textbf{Few-shot Prompting~\cite{brown2020languagemodelsfewshotlearners}}. This technique often involves prepending identical instructions or examples (e.g., demonstrations of tool usage) to various distinct prompts.
    \item \textbf{Self-consistency~\cite{wang2023selfconsistencyimproveschainthought}}. It uses a standard Chain-of-Thought (CoT) few-shot prompt, and samples a diverse set of reasoning paths. This initial CoT prompt acts as the shared prefix for multiple sampling iterations.
    \item \textbf{Tree-of-thoughts~\cite{yao2023treethoughtsdeliberateproblem}.} It explores multiple solution paths by building a tree of intermediate steps, where each branch represents a possible reasoning path. Its tree structure allows prefix sharing, as sibling nodes reuse common parent computations.
    \item \textbf{Speculative Decoding~\cite{specinfer}.} Within the verification phase of speculative decoding, the generation process can form tree-structured queries where nodes representing sequential tokens share common ancestor paths, enabling prefix sharing.
\end{itemize}

Existing research~\cite{zheng2024sglangefficientexecutionstructured, qin2024mooncakekvcachecentricdisaggregatedarchitecture, hu2024memservecontextcachingdisaggregated,zheng2024batchllmoptimizinglargebatched, zhao2024blendserveoptimizingofflineinference,wang2025echoefficientcoschedulinghybrid} leverages KV cache reuse for requests sharing the same prompt prefixes, thereby accelerating the prefill phase and reducing memory consumption. They typically maintain the KV cache in a tree fashion, where each node corresponds to a prefix. However, when accessing the KV cache, the system still assumes \emph{a logical 4D tensor structure for the Key and Value tensors, thus still suffering from duplicated global memory accesses}.

These observations surface two key challenges in prefix-shared decoding---(i) kernel design over an irregular KV-cache forest and (ii) workload imbalance under highly skewed per-node workloads---which we address in Section~\ref{sec:kernel} and Section~\ref{sec:wb}, respectively.

\section{Overview}
\label{sec:overview}

\subsection{Requirement}
To achieve efficient prefix-shared decoding, the developed kernel should satisfy the following requirements:

\stitle{IO Efficient:} As the decode stage is memory-intensive, the prefix-shared decoding kernel should be able to minimize the global memory access overhead by leveraging the shared memory to conduct the attention computation between the KV cache of shared prefix tokens and the query tensor of these requests. In addition, the developed kernel should have sufficient parallelism for inter-block parallelism and only introduce limited extra overhead for synchronization and reduction.

\stitle{Workload Balance:} As the computation is shifted from regular 4D tensor to tree of 3D tensors, the workload distribution among different blocks is unbalanced. The developed kernel should be able to balance the workload among different blocks, to put it in another way, the divided subtasks should have a similar workload.

\subsection{System Architecture}

\begin{figure}[tb]
    \centering
    \includegraphics[width=0.9\linewidth]{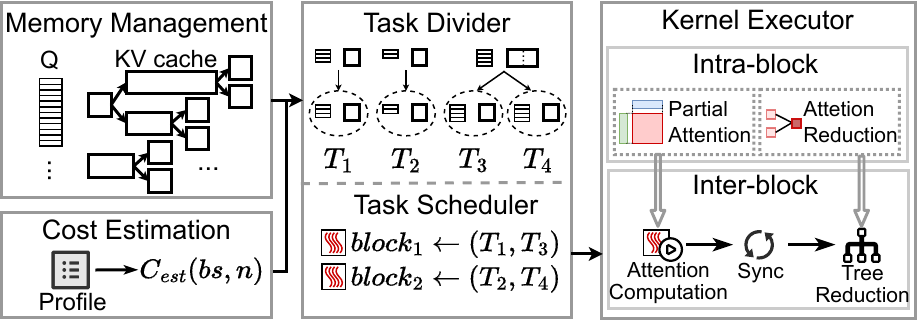}
    \vspace{-.5em}
    \caption{Overview of \name.
    }
    \vspace{-.5em}
    \label{fig:overview}
\end{figure}

\stitle{Memory Manager (Section~\ref{subsec:kv_cache_management}):} The KV cache of the current running batch is materialized as a tree of tensors in the global memory, where each node maintains the KV cache of a chunk of tokens shared by multiple requests or owned by a single request.
The queries of the requests are materialized as a query tensor.
Moreover, we maintain a lightweight index that links each query to the relevant nodes in the KV-cache tree, enabling the kernel to directly locate and coalesce reads to shared-prefix KV blocks.

\stitle{Kernel Executor (Section~\ref{sec:intra-block kernel primitive},~\ref{subsec:inter-block_and_tree-reduction}):} The kernel executor exposes both intra-block and inter-block parallelism.
At the intra-block level, we reuse and adapt two primitives---partial attention computation (PAC) and partial output reduction (POR)---to efficiently compute and aggregate partial results.
At the inter-block level, \name schedules PAC across the prefix-tree blocks and then performs a parallel, tree-structured reduction based on POR to merge partial outputs per query, avoiding the overhead of launching many small, sequential reduction kernels.

\stitle{Cost Estimator (Section~\ref{sec:cost estimator}):} The cost estimator predicts the execution time of PAC for a given (query, KV-block) shape.
Because PAC may be either compute-bound or memory-bound depending on tensor shapes and GPU characteristics, \name uses lightweight micro-benchmark profiling on the target GPU/model and interpolates to estimate costs for unprofiled workloads.

\stitle{Task Divider and Scheduler (Section~\ref{subsec:task_division_scheduling}):} Given a decoding batch, the workload across prefix-tree nodes is highly skewed.
\name uses the cost estimator to divide and assign work with a global view of the prefix tree, aiming to balance block-level execution time while avoiding overly fine-grained partitioning that would increase scheduling and reduction overhead.

\section{GPU Kernel Design} \label{sec:kernel}

\rOneFC{In FlashAttention/FlashDecoding, KV tensors follow a regular 4D layout that naturally supports efficient materialization, block partitioning, and reduction.}
\rOneFC{In prefix-shared decoding, the KV cache becomes a forest of per-prefix 3D tensors, which complicates KV/query indexing, increases task-division complexity, and requires tree-structured reduction to merge partial results.}

We first introduce our compute-centric KV cache management for facilitating the attention computation in the prefix-shared decoding kernel.
Then, we formulate two essential intra-block kernel primitives:
the partial attention computation kernel and the partial output reduction kernel, which serve as the building blocks of our prefix-shared decoding kernel.
Finally, we introduce the inter-block kernel executor, which orchestrates the parallel execution of these intra-block kernel primitives to maximize computational efficiency.

\begin{figure}[H]
	\centering
	\includegraphics[width=0.7\linewidth]{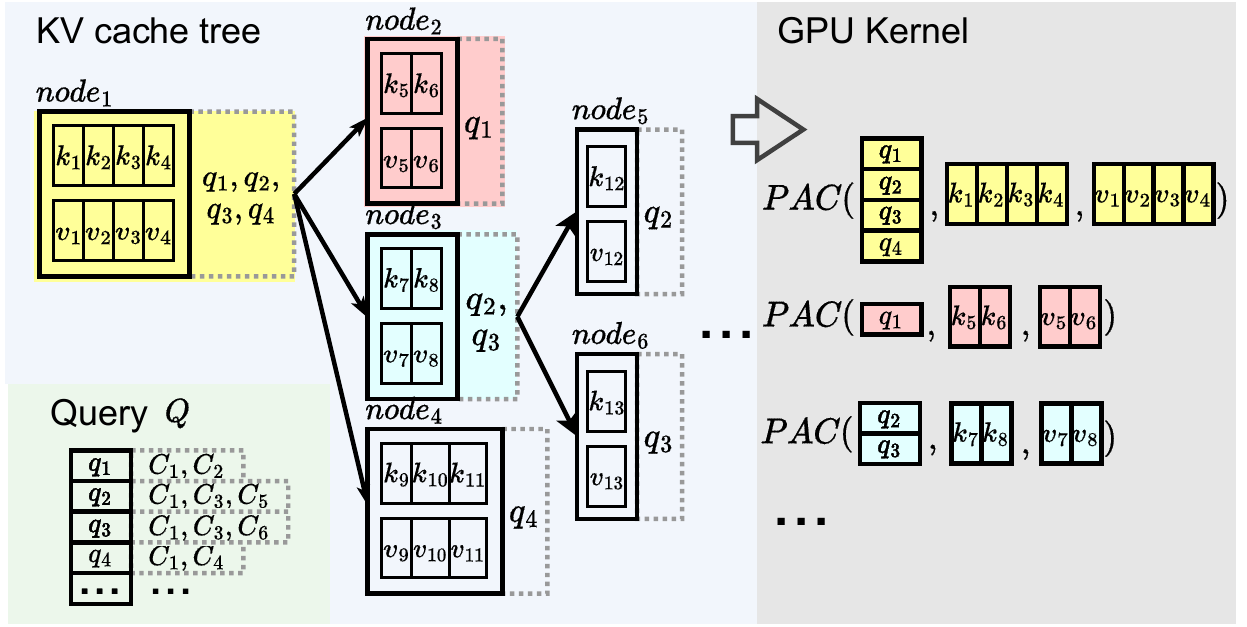}
	\caption{KV cache forest.}
	\label{fig:kv cache forest}
\end{figure}

\subsection{KV Cache Management} \label{subsec:kv_cache_management}
Different from the traditional KV cache management systems, e.g., PagedAttention~\cite{vllm_sosp23} targeting for maintaining the KV cache in the GPU memory, our prefix-shared decoding kernel has further responsibilities to support the prefix-shared decoding, which requires efficient indexing for the further partial attention computation and reduction operations.

\stitle{Formal model and notation.}
Let $\mathcal{R}=\{r_1,\ldots,r_B\}$ be the batch of requests. Each request $r$ has a current query vector $q_r\in\mathbb{R}^{d_k}$, and a (possibly shared) prefix token sequence $P(r)=(t_1,\ldots,t_{\ell(r)})$.
We maintain a forest (with a virtual root) $\mathcal{F}=(\mathcal{N},\mathcal{E})$ of KV-cache nodes, where each $n\in\mathcal{N}$ stores tensors
\[
	K_n\in\mathbb{R}^{|n|\times d_k},\qquad V_n\in\mathbb{R}^{|n|\times d_v},
\]
and an ordered token index set $\mathrm{Tok}(n)=\{1,\ldots,|n|\}$.
An edge $(p\to c)\in\mathcal{E}$ encodes $p$ is a prefix of $c$ (write $p\prec c$). A virtual root $\varnothing$ connects all prefix roots.

Define the \emph{prefix path} of $r$:
\[
	\begin{aligned}
		             & \pi(r)=(n_1,\ldots,n_{m(r)})\quad                            \\
		\text{with}  & \quad \varnothing=n_0\prec n_1\prec\cdots\prec n_{m(r)}\quad \\
		\text{s.t. } & \quad \text{concat}\big(\{K_{n_i}\}\big)\text{ matches }P(r)
	\end{aligned}
\]
All queries are stacked into $Q\in\mathbb{R}^{B\times d_k}$ with $Q[r,:]=q_r^\top$.
Let the global token address space be
\[
	\Omega=\{(n,i)\mid n\in\mathcal{N},\, i\in\mathrm{Tok}(n)\},
\]
and a bijection (logical flattening) $\kappa:\Omega\overset{\sim}{\to}\{1,\ldots,L_{\mathrm{tot}}\}$ that orders tokens across nodes; this induces logical tensors $K,V$ by
\[
	K[\kappa(n,i),:]=K_n[i,:],\qquad V[\kappa(n,i),:]=V_n[i,:].
\]
(We never physically materialize $K,V$ monolithically; $\kappa$ is index-level.)

\stitle{Tree-based KV cache management.}
As shown in Figure~\ref{fig:kv cache forest}, we manage the KV cache as a tree of tensors, where each node in the tree represents a chunk of KV cache, and the edge between the father and child nodes represents the relationship between the two chunks, where the father node is the prefix of the child node.
Moreover, the queries from all requests are consolidated into a single query tensor, with each row corresponding to an individual request's query.
Notice, Figure~\ref{fig:kv cache forest} illustrates the example of all requests sharing the same prefix in node 1, while in practice, two or more prefixes may be shared by different requests. Therefore, we introduce a virtual root node to represent the root of the tree, which connects all the prefixes of the requests. This virtual root node allows batching different prefixed requests together, which allows the kernel to support even the non-prefix-shared decoding.

In addition to the tensor enclosed in solid lines, we also maintain the following index structures enclosed in dashed lines, which record the bijective mapping between the KV cache and the query tensor.

\stitle{Tensors preparation for partial attention computation.}
As shown in the right part of Figure~\ref{fig:kv cache forest}, the queries shared the same prefix maintained in the KV cache node actually form a partial attention computation. Therefore, for each KV cache node, we need to maintain the \emph{set of queries} of the requests that share the same prefix with this node, which facilitates the aggregation of the queries that share the same prefix, allowing further partial attention computation between tensors formed by the queries and the corresponding KV cache node.
Instead of preparing the query tensor in the global memory,
the query tensor can be aggregated in thread block shared memory during partial attention kernel launch, reducing memory overhead.

\paragraph{Formal per-node assembly.}
For node $n$, define the per-node query tensor
\[
	Q^{(n)}\stackrel{\mathrm{def}}{=}Q[I_n,:]\in\mathbb{R}^{|\mathcal{R}(n)|\times d_k}.
\]
We assemble $Q^{(n)}$ on-chip (shared memory) at kernel launch:
\[
	Q^{(n)}[a,:]\leftarrow Q[i_a,:]\quad (a=1,\ldots,|I_n|),
\]
and stream tiles of $(K_n,V_n)$ to shared memory for the GEMM and value-weighted reduction.

\stitle{Indexing of partial results for each query.}
As introduced in Section~\ref{sec:attention background}, the softmax operation inherently works globally across each row of the attention score tensor, essentially operating at the query level.
This characteristic introduces a specific indexing requirement: for each query, the system needs to maintain a record of which KV cache nodes constitute its prefix.
This query-specific tracking enables correct indexing and retrieval of partial attention computation results, forming a critical component when aggregating across multiple partial computations.

\paragraph{Streaming softmax across nodes.}
Our goal is to compute attention for each query $r\in\mathcal{R}$ without materializing the global score vector: we stream over nodes (blocks) and accumulate the softmax numerator and denominator on the fly. Let $J_r$ be the set of nodes visible to $r$. When visiting a node $n\in J_r$, we compute the scaled scores
\[
	s_{r,n}=\frac{1}{\sqrt{d_k}}\,Q[r,:]\,K_n^\top\in\mathbb{R}^{|n|},
\]
and apply the visibility mask to obtain
\[
	\tilde{s}_{r,n}[j]=
	\begin{cases}
		s_{r,n}[j], & \text{if $(n,j)$ is visible to } r, \\
		-\infty,    & \text{otherwise}.
	\end{cases}
\]
We then perform \emph{local} stabilization within the node: with the nodewise maximum
$m_{r,n}=\max_j \tilde{s}_{r,n}[j]$, define
\[
	d_{r,n}=\sum_j e^{\tilde{s}_{r,n}[j]-m_{r,n}},\qquad
	u_{r,n}=\sum_j e^{\tilde{s}_{r,n}[j]-m_{r,n}}\,V_n[j,:]\in\mathbb{R}^{d_v}.
\]
Here $d_{r,n}$ is the node’s contribution to the softmax denominator, and $u_{r,n}$ is the vector contribution to the numerator. Masked (invisible) positions are set to $-\infty$, hence contribute $0$ after exponentiation.

To avoid overflow/underflow when merging across nodes, we maintain \emph{per-query accumulators}
\[
	M[r]\leftarrow -\infty,\quad D[r]\leftarrow 0,\quad U[r,:]\leftarrow \mathbf{0}\in\mathbb{R}^{d_v}.
\]
When incorporating node $n$, we use a common log-sum-exp reference
$\hat{M}=\max\big(M[r],\,m_{r,n}\big)$ and combine the old and new contributions in the same exponential frame.
\[
	M[r]\leftarrow \hat{M}.
\]
Intuitively, $M[r]$ tracks the running global maximum logit, while $D[r]$ and $U[r,:]$ are the denominator and numerator evaluated relative to that maximum. Iterating over $n\in J_r$ in this way preserves numerical stability and requires memory only proportional to the current node.

\subsection{Intra-Block Kernel Primitive}
\label{sec:intra-block kernel primitive}
Following the KV cache management, we abstract two necessary operations, computing the partial attention within each KV cache node and reducing the partial attention computation results between the KV cache nodes, into two intra-block kernel primitives, namely, the partial attention computation (PAC) kernel and the partial output reduction (POR) kernel.
As the GPU architecture suggests (as shown in Figure~\ref{fig:intropic}(b)), the intra-block kernel primitives are executed upon the thread blocks configured with on-chip shared memory, our intra-block kernel primitives are designed to be executed in shared memory, which can significantly reduce the global memory access overhead.

\begin{figure}[H]
	\vspace{-1em}
	\begin{minipage}{\linewidth}
		\begin{minipage}{0.48\columnwidth}
			\begin{algorithm}[H]
				\caption{PAC}
				\label{alg:pac}
				\begin{algorithmic}[1]
					\State \textbf{Input:} $Q$, $K$, $V$
					\State \textbf{Output:} $O$
					\State $S \gets Q \cdot K^T / \sqrt{d}$
					\State $S \gets \text{SOFTMAX}(S)$
					\State $O \gets S \cdot V$
					\State \textbf{return} $O$
				\end{algorithmic}
			\end{algorithm}
		\end{minipage}
		\begin{minipage}{0.5\columnwidth}
			\begin{algorithm}[H]
				\caption{POR}
				\label{alg:por}
				\begin{algorithmic}[1]
					\State \textbf{Input:} $O_1$, $O_2$, $m_1$, $m_2$, $s_1$, $s_2$
					\State \textbf{Output:} $O$
					\State $m \gets \max(m_1, m_2)$
					\State $s \gets s_1  e^{m_1 - m} + s_2 \cdot e^{m_2 - m}$
					\State $O \gets \frac{O_1\cdot s_1 \cdot e^{m_1 - m}+ O_2\cdot s_2 \cdot e^{m_2 - m}}{s}$
					\State \textbf{return} $O$
				\end{algorithmic}
			\end{algorithm}
		\end{minipage}
	\end{minipage}
\end{figure}

\stitle{Partial attention computation (PAC) kernel:}
As the name suggests, the partial attention computation kernel is responsible for performing attention computation between a query sub-tensor ($Q\in \mathbb{R}^{n_q \times d}$) and its corresponding KV cache sub-tensor ($K, V \in \mathbb{R}^{n \times d}$), where $n_q$ queries in the query sub-tensor share the same prefix with length $n$ maintained in the KV cache sub-tensor.

We observe that the computation of the PAC kernel is exactly the same as the computation of the attention operation, except that the input query tensor in the PAC kernel is sourced from multiple requests while the input query tensor in the ordinary attention operation are sourced from different tokens of the same request.
We efficiently support the intra-block partial attention computation as demonstrated in Algorithm~\ref{alg:pac}, requiring only minimal modifications to the FlashAttention kernel.
Instead of limiting the memory consumption of query, key, and value tensors to shared memory, our intra-block kernel primitives further partition the partial attention computation to leverage the shared memory, and sequentially process partitioned computations, thereby supporting larger workloads while maintaining the memory efficiency by leveraging the shared memory.

To further reduce the memory access overhead, we also optimize the data loading pattern of the PAC kernel to better leverage the shared memory. Specifically, instead of loading the key and value tensors from the global memory for each partial attention computation, we load the key and value tensors into the shared memory once, and then reuse them for multiple query.
This optimization can significantly reduce the memory access overhead, especially when the ratio of GQA is high.
Please note that this optimization also reduces the inefficiency inside Tensor Cores, as the padding of the query tensor is reduced.

\stitle{Partial output reduction (POR) kernel:}
As the operation between a query and its corresponding KV cache will be partitioned into several partial attention computations as the KV cache is divided into several KV cache nodes, we need to merge the results of these partial attention computations to obtain the final output of the query. Instead of reducing the whole results, the POR kernel is a binary reduction operation, which merges two partial attention computation results $O_1\in \mathbb{R}^{n_q \times d}$ and $O_2\in \mathbb{R}^{n_q \times d}$ sourced from different KV cache nodes of the same query set.

As shown in Algorithm~\ref{alg:por}, the POR kernel takes $O_1$, $O_2$, and their corresponding max attention scores $m_1$, $m_2$ and the sum of exp attention scores $s_1$, $s_2$ as input, and outputs the final output $O$ of the query. Line 3 computes the maximum attention score $m$ of the two partial attention computation results while line 4 computes the sum of the attention scores $s$ of the two partial attention computation results, both of which are used to normalize the final output $O$. Subsequently, line 5 renormalizes the two partial attention computation results $O_1$ and $O_2$ and merges them into the final output $O$.
As the size of the output $O$ can be easily fitted into the shared memory, POR kernel by default will be executed in the shared memory.

\subsection{Inter-Block Launching and Tree-Reduction} \label{subsec:inter-block_and_tree-reduction}

On top of the intra-block kernel primitives, we develop the inter-block kernel executor, which is responsible for executing the intra-block kernel primitives in a parallel manner to conduct the prefix-shared attention computation for a batch of requests.

\begin{algorithm}[ht]
	\caption{\name}
	\label{alg:our method}
	\begin{algorithmic}[1]
		\State \textbf{Input:} $\mathbf{Q}, (\mathbf{K}, \mathbf{V})$
		\State \textbf{Output:} $\mathbf{O}$
		\State Initialize $\mathbf{O}_{tree}$ with the tree structure of $(\mathbf{K}, \mathbf{V})$
		\For{$(K, V) \in (\mathbf{K}, \mathbf{V})$}
		\State Aggregate the tensor $Q$ from query index in $(K, V)$
		\State $\mathbf{O}_{tree}[(K, V).index] \gets PAC(Q, K, V)$
		\EndFor
		\For{$O\in\mathbf{O}_{tree}$}
		\State $\mathbf{O}[O.query\_index] \gets POR(\mathbf{O}[O.query\_index],O)$
		\EndFor
		\State \textbf{return} $\mathbf{O}$
	\end{algorithmic}
\end{algorithm}

Algorithm~\ref{alg:our method} illustrates how to sequentially launch the intra-block kernel primitives to perform the attention computation. It mainly consists of two steps: 1) launching the PAC kernel for each KV cache node (lines 4-6), 2) conducting the tree reduction to merge the results for each query (lines 7-8).

The PAC kernel launching is quite straightforward, as each KV cache node and its corresponding query tensor can be easily indexed by the KV cache management system. Given the computations of the PAC kernel are independent, we can leverage embarrassingly parallelism in line 4 to launch the PAC kernel for each KV cache node in parallel. Moreover, a synchronization is conducted to ensure that all partial attention computation results are prepared before conducting the tree reduction operation.

As shown in line 6, the partial attention computation results of the PAC kernel are stored in a tree structure same as the KV cache management system. Therefore, the reduction operation should be conducted in a tree structure, which introduces the challenge in parallelization.

\stitle{Parallelization of the tree reduction operation.}
We notice that the reduction operation satisfies the \emph{associative and commutative properties}.
Specifically, recalling Algorithm~\ref{alg:por}, the reduction operation between two partial attention computation results $O_1$ and $O_2$ is independent of the order of the reduction operation, i.e., associative ($POR(O_1, O_2) = POR(O_2, O_1)$) and commutative ($POR(POR(O_1, O_2), O_3) = POR(O_1, POR(O_2, O_3))$).
These properties allow us to reorganize the reduction operation order, which facilitates the parallelization.
Moreover, lines~7-8 in Algorithm~\ref{alg:our method} implicitly indicate that the \emph{reduction operation of different queries is independent}, no matter how the tree is structured. These two observations suggest that we can transform the tree reduction operation into $bs$ independent series of reduction operations, where $bs$ is the number of queries in the batch. Moreover, the reduction operation of non-adjacent edges in the series of each query can be conducted in parallel, as the commutative property of the reduction operation allows us to change the order of the reduction operation.
Hence, we can easily speed up the tree reduction by exploiting parallelism in two dimensions: 1) parallelizing the reduction operation of different queries, 2) parallelizing the reduction operation of different nodes in the tree by replicating $\mathbf{O}$ and conducting an addition reduction on the replicated $\mathbf{O}$.

\stitle{Complexity analysis.}
The IO complexity of \name can be denoted as $O(h\cdot d\sum_{i=1}^{node\_num} n[i])$ while the IO complexity of FlashAttention is $O(h\cdot d\sum_{i=1}^{node\_num} n[i]\times n_q[i])$, where we ignore the cost of loading the query tensor and writing the output tensor, as it is negligible compared to the cost of loading the KV cache tensor (i.e., $n_q[i]\ll n[i]$). Intuitively, given $\overline{n_q}$ as the weighted average of the shared ratio of the KV cache, i.e.,
$\overline{n_q}=\frac{\sum_{i=1}^{node\_num} n[i]\times n_q[i]}{\sum_{i=1}^{node\_num} n[i]}$
the IO complexity of \name is about $\overline{n_q}$ times lower than that of FlashAttention. Regarding the computation complexity, \name is the same as FlashAttention.

\section{Workload Balance} \label{sec:wb}

\rOneFC{For a KV-cache node, both the KV length (prefix length) and the number of associated queries (degree of sharing) can vary widely across nodes.}
\rOneFC{As a result, naively launching one PAC kernel per node leads to severe inter-block load imbalance and GPU under-utilization; however, overly fine-grained splitting increases scheduling and reduction overhead and may underutilize tensor cores within each block.}

In this section, we first formulate the optimization problem of task division and scheduling, which is np-hard, and propose a heuristic solution through pruning. Recognizing the inaccurate of theoretical cost estimation, we further propose a profile-based estimator.

\subsection{Task Division and Scheduling}
\label{subsec:task_division_scheduling}

An intuitive approach is to further divide the KV cache node into several sub-nodes as well as the queries into several sub-queries, and then assign each sub-node to a thread block.
However, determining the granularity of this division presents significant challenges.
On the one hand, fine-grained task division can achieve better workload balance, but it may result in a large number of tasks, which will introduce additional scheduling and reduction overhead, and a fine-grained task can also lead to resource under-utilization within each thread block due to insufficient workload for tensor core in each block.
On the other hand, coarse-grained task divisions may still suffer from workload imbalance, which also leads to under-utilization of GPU resources due to inter-block stallings.

Therefore, we formulate the task division and scheduling problem as an optimization problem, where we aim to find the optimal task division and scheduling strategy to minimize the execution time of the slowest thread block.

\stitle{Division and scheduling formulation.} The partial attention computations can be modeled as a set of tasks, where each task is a tuple $(Q[i], K[i], V[i])$. As the dimension of feature $d$ is fixed, we can ignore it in the task division formulation, thereby we use $\mathbf{T}[i]=(n_q[i], n[i])$ to represent $i$-th task, where $n_q[i]$ is the number of queries in the $i$-th task and $n[i]$ is the sequence length of the KV cache node in the $i$-th task. We use $t$ to denote the number of KV cache nodes, i.e., the number of tasks.
For each task $\mathbf{T}[i]$, we can divide it both horizontally (i.e., in the query dimension, whose number of horizontal slices denotes $b_q[i]$) and vertically (i.e., in the KV cache dimension, whose number of vertical slices denotes $b_k[i]$). Therefore, we aim to find the optimal task division strategy $\{(b_q[i], b_k[i])\}_{i=1}^t$ to minimize the total execution time of all tasks.

In addition to task division, we also need to consider task scheduling or task assignment strategy. Assuming we have $m$ thread blocks, we need to assign the tasks to the thread blocks. The task assignment strategy can be represented as a tensor $\mathbf{A}\in \mathbb{N}^{m\times t}$, where $\mathbf{A}[i,j]$ is the number of divided sub-tasks of task $j$ assigned to thread block $i$, $m$ is the number of thread blocks.
To facilitate our discussion, we use $\mathbf{C}[j]$ to represent the estimated execution time of a sub-task of task $j$, which can be computed by the cost estimator will be introduced in Section~\ref{sec:cost estimator}.

Formally, we can formulate the task division and scheduling problem as follows:
\begin{equation}
    \label{eq:optimization}
    \begin{aligned}
         & \mathop{\arg\min}_{b_q,b_n, \mathbf{A}} & Cost=\max_{i=1}^m(\sum_{j=1}^t \mathbf{C}[j]*\mathbf{A}[i,j]),           \\
         & s.t.                                    & \forall j\in [1, t], \sum_{i=1}^m \mathbf{A}[i,j] = b_q[j] \cdot b_k[j]. \\
    \end{aligned}
\end{equation}
The objective function is to minimize the maximum execution time of all thread blocks, where the cost of each thread block is the sum of the execution time of all subtasks assigned to this thread block. The constraint is to ensure that all subtasks of each task are assigned to the thread blocks.

\stitle{Solver.} The above problem is an advanced parallel task scheduling problem~\cite{Boundsforcertainmultiprocessinganomalies}, which is NP-hard.
Subsequently, we first simplify the problem, and then narrow down the search space by obtaining the lower and upper bounds of the $cost$, and finally exhaustively search the task division and scheduling strategy.

We observe that the $n_q\ll n$ in most cases as the KV cache is usually much larger than the number of queries. If we divide the task into the query dimension, the cost increases significantly, as it actually misses the opportunity to combine the KV cache memory access. Therefore, we set the number of horizontal slices $b_q[i]$ to 1, and focus on the vertical slices $b_k[i]$.

To further narrow down the search space, we easily find the following two properties inequalities.
1) Noticing that the sum cost of the sub-tasks is no less than the cost of the original task, and the max cost of each block is no less than the average cost of all blocks, we can reach the following inequality:
\begin{equation}
    \label{eq:inequality of average}
    Cost\geq \frac{1}{m}\sum_{i=1}^m(\sum_{j=1}^t \mathbf{C}[j]*\mathbf{A}[i,j])
\end{equation}
Moreover, with more fine-grained task division, the average cost of all blocks will be larger, as the workload is not reduced, but the scheduling overhead is increased. This monotonicity and the inequality in Equation~\ref{eq:inequality of average} can be used to determine the lower bound of the cost (denoted as $cost_l$) through binary search.

Therefore, we can narrow down the search space by setting the upper bound of the division number of each KV cache node as
\begin{equation}
    \label{eq:inequality of upper}
    b_k[i]\leq \lceil{\frac{C_{est}(n_q[i], n[i])}{cost_l}}\rceil,
\end{equation}
where $C_{est}(n_q[i], n[i])$ is the estimated execution time defined in Section~\ref{sec:cost estimator}. This inequality restricts the further division of the KV cache node when the cost is lower than the average cost, as under such conditions, subtasks are enough to saturate the GPU's block-level parallelism, while further division will lead to more overhead.

In practice, the equation~\ref{eq:inequality of upper} sets the division number of most tasks to 1, whose workload is significantly smaller than the average cost. For example, in documented question-answering tasks, despite the shared document KV cache node ($n\approx 10k$), the workload of the question KV cache node for each request ($n\approx 50$) is usually much smaller.
Therefore, we gird search the division number of each KV cache node and choose the optimal division.

\subsection{Cost Estimation}
\label{sec:cost estimator}

We observe that \emph{the execution cost of partial attention computation varies from the theoretical result.} It is easy to compute the theoretical workload of the partial attention computation given $Q\in \mathbb{R}^{n_q\times d}$ and $K,V\in \mathbb{R}^{n\times d}$. The computation mainly involves two matrix multiplications, i.e., $QK^T$ and $AV$, where $A$ is the attention score tensor. The theoretical workload of the first matrix multiplication is $O(n_q\times n\times d)$, and the second matrix multiplication is $O(n_q\times n\times d)$, resulting in a total theoretical workload of $O\left(n_q\times n\times d\right)$. Moreover, the global memory access is the sum of the memory access of $Q$, $K$ and $V$, which is $O\left((n_q+2n)\times d\right)$. The divergence between the computation and memory access makes the execution cost hard to estimate. Moreover, the kernel has a constant launch overhead, which is independent of the workload. Therefore, for the small workload, the execution cost is dominated by the kernel launch overhead, while for the large workload, the execution cost is dominated by the computation and memory access.

As revealed in Table~\ref{tab:block_execution_time}, different shapes of the tensors will lead to different execution costs due to the varying hardware resource utilization. When the workload is small, the execution time is dominated by the kernel launch overhead. For small $n_q$ and large $n$, the PAC kernel is memory bound, thus cost scales almost linearly with $n$, while for large $n_q$ and $n$, the PAC kernel is compute bound.

\begin{table}[t]
    \caption{Thread block execution time (ms), $d=128$.}
    \centering
    \scalebox{0.87}{
        \begin{tabular}{|l|c|c|c|c|c|c|c|}
            \hline

            \diagbox{$n$}{$n_q$} & \multicolumn{1}{c|}{1} & \multicolumn{1}{c|}{2} & \multicolumn{1}{c|}{5} & \multicolumn{1}{c|}{10} & \multicolumn{1}{c|}{20} & \multicolumn{1}{c|}{50} & \multicolumn{1}{c|}{100} \\
            \hline
            512                  & 0.036                  & 0.035                  & 0.036                  & 0.043                   & 0.048                   & 0.074                   & 0.112                    \\
            1,024                & 0.043                  & 0.043                  & 0.044                  & 0.054                   & 0.062                   & 0.109                   & 0.122                    \\
            2,048                & 0.060                  & 0.059                  & 0.059                  & 0.079                   & 0.094                   & 0.124                   & 0.145                    \\
            4,096                & 0.092                  & 0.092                  & 0.093                  & 0.126                   & 0.147                   & 0.156                   & 0.183                    \\
            8,192                & 0.156                  & 0.157                  & 0.156                  & 0.199                   & 0.189                   & 0.195                   & 0.266                    \\
            16,384               & 0.283                  & 0.282                  & 0.283                  & 0.301                   & 0.303                   & 0.471                   & 0.746                    \\

        \hline
    \end{tabular}
    }
    \label{tab:block_execution_time}
\end{table}

Therefore, we propose a profile-based approach to estimate the execution cost of the partial attention computation between each KV cache node and its corresponding queries.
With a given hardware configuration and a given model (i.e., the dimension of the feature $d$), we observe that only $n$ and $n_q$ are the two parameters that affect the execution time of the PAC kernel. Therefore, before deploying the model, we can profile the PAC kernel with various sizes of the KV cache node $n$ and various numbers of queries $n_q$, and record the execution time. For the unprofiled computation, we use interpolation to estimate the execution time.
After profiling, we have the following cost estimation function:
\begin{equation}
    C_{est}(n_q, n),
\end{equation}
which is the estimated execution time of the partial attention computation between a KV cache node with sequence length $n$ and $n_q$ queries.

Recall the cost in Equation~\ref{eq:optimization}, for the cost of a subtask of task $T[i]$, we can estimate the execution time as:
\begin{equation}
    \mathbf{C}[j] = C_{est}(\frac{bs_j}{v_j}, \frac{n_j}{h_j}).
\end{equation}

\section{Implementation}
\label{sec:implementation}

We implement the \name kernel module in CUDA/C++ on top of NVIDIA CUTLASS (about 1{,}700 LOC) and the task-division module in C++ (about 300 LOC).
The implementation is modular to facilitate extension to new models and attention variants.
\rAllFC{The kernel supports widely used attention layouts in modern LLMs, including MHA, MQA, and GQA.}

\stitle{CoDec Kernel Module.}
The \name kernel takes the query tensor, paged KV cache, and decoding metadata (e.g., block tables and context lengths) as inputs and writes attention outputs to the designated output tensor.
\rTwoFC{It follows the same paged KV-cache layout as PagedAttention~\cite{vllm_sosp23} and exposes an attention interface compatible with FlashDecoding, making integration into vLLM straightforward (i.e., swapping the attention backend while reusing the existing batching and memory manager).}
The kernel is implemented with CUTLASS to leverage fine-grained pipelining across memory and compute.

\rAllFC{For models using multi-head latent attention (MLA), \name can be extended by first reconstructing per-head KV blocks from the latent representation and then applying the same prefix-aware attention and reduction pipeline.}

\stitle{CoDec Task Division Module.}
To maximize the efficiency of the task division module, we implement it as a C++ module that interfaces with the main decoding loop. This module monitors the decoding process and dynamically divides tasks based on the current workload and resource availability.
To reduce overhead, we perform task division every few decoding steps rather than at every step.
\rAllFC{Empirically, the partitioning overhead accounts for 1.3\%--2.5\% of the total attention time in our evaluation, and the parallel reduction contributes less than 10\% overhead relative to partial attention computation under typical shared-prefix workloads.}

\section{Evaluation} \label{sec:eval}
In this section, we evaluate the performance of \name on various tasks. In summary, we want to answer the following questions:
\begin{itemize}
    \item \textbf{Section~\ref{subsec:sota}: What is the benefit of using \name?} We compare \name with the SOTA attention kernel, FlashAttention~\cite{dao2023flashattention2} and FlashDecoding~\cite{flashdecoding}, in terms of attention operation time, end-to-end time, and global memory IO.
    \item \textbf{Section~\ref{subsec:ablation}: How does each optimization contribute to the performance?} We conduct an ablation study to analyze the contribution of each optimization in \name.
    \item \textbf{Section~\ref{subsec:division-granularity}: How do the specific design choices impact \name's performance?} We analyze the impact of key design decisions in \name, particularly focusing on the optimal division granularity for different sequence lengths.
    \item \textbf{Section~\ref{subsec:task-division-overhead}: What is the overhead of task division?} \rAllFC{We quantify the CPU overhead of computing a division plan and show how it scales with batch size.}
    \item \textbf{Section~\ref{subsec:gpu-varing}: How does \name perform on different GPUs?} We evaluate the performance of \name on different GPUs, including NVIDIA H800, A100, RTX 4090, A30, and RTX A6000.
\end{itemize}

\subsection{Experimental Setup}
\label{subsec:setup}
\rTwoFC{\name is implemented in around 2{,}000 lines of CUDA/C++.}
\rTwoFC{By default, we use Qwen3-4B, which has 32 query heads, 8 key/value heads, and head dimension 128.}
\rTwoFC{Unless otherwise specified, we run experiments on a single NVIDIA A100 GPU (40GB, PCIe) with CUDA Toolkit 11.8 (runtime version 12.2), vLLM 0.6.6, and Python 3.10.
    We report the mean over 3 runs.}
\rAllFC{To study generality, we additionally evaluate across attention layouts (MHA/MQA/GQA) and multiple model families/sizes (Section~\ref{subsec:attention-variants}).}

\stitle{Workloads.}\rTwoFC{We evaluate both controlled synthetic prefix trees and a real-world long-context dataset.
    For synthetic workloads, we use controlled prefix-sharing trees; the detailed workload specifications are described in Section~\ref{subsec:sota}.
    For real-world evaluation, we use the LooGLE dataset and materialize shared prefixes by grouping queries that share the same document context.}

\stitle{Baselines and metrics.} \rTwoFC{We compare against FlashDecoding~\cite{flashdecoding} (decode-stage attention kernel) and FlashAttention~\cite{dao2023flashattention2} where applicable.} \rThreeFC{We also include FlashInfer's multilevel cascade attention~\cite{cascade-inference} in a complementary shared-prefix throughput comparison (Figure~\ref{fig:dataset-e2e}).}
\rTwoFC{We report attention-kernel time (CUDA events), end-to-end decoding time per output token (TPOT) in comparison with vLLM, and global memory traffic of attention kernels measured via GPU profiling counters.}
\rAllFC{Finally, we discuss the overhead of \name's workload partitioning and tree reduction: while the benefit naturally diminishes when prefix sharing is limited, the partitioning overhead remains small in practice because we amortize partitioning by reusing a division plan for multiple decoding steps (Section~\ref{sec:implementation}).}

\subsection{Comparison with SOTA}
\label{subsec:sota}

To further reveal the performance characteristics of \name, we conduct a series of experiments to evaluate the impact of different workloads on the performance of \name. We consider the following workloads:

\stitle{Workload.} By default, we consider the 2-level tree structure, where the root node is the prefix shared by all requests, and the leaf nodes are the KV cache of each request. This is a common case in document QA tasks, where all requests share the same document. Generally, we consider the following workloads:
\begin{itemize}
    \item \textbf{Varying sequence length:} Fixing the full binary shared prefix tree with depth 2. We vary the sequence length of the non-shared prefix from 512 to 8,192 tokens.
    \item \textbf{Varying batch size:} Fixing the full binary shared prefix tree with depth 2 and root node context length of 120k, we vary the batch size, i.e., the number of requests.
    \item \textbf{Varying tree depth:} In this workload, we choose the full binary tree structure, where each node has two children, and vary the tree depth from 2 to 6.
    \item \textbf{Varying shared prefix ratio:} We vary the shared prefix ratio by controlling the number of shared tokens in the default 2-level tree structure with a total context length of 120k. The shared prefix ratio is defined as the number of shared tokens divided by the total number of tokens in the KV cache tree.
    \item \textbf{Varying tree shape:} We consider 1) binary tree (2T), 2) ternary tree (3T), 3) quaternary tree (4T), 4) quinary tree (5T), each with the same workload. Moreover, we also consider 5) degenerate tree (DT), where only the left nodes have children.
\end{itemize}

\begin{figure*}[tp]
    \centering
    \begin{subfigure}{0.195\linewidth}
        \centering
        \includegraphics[width=\linewidth]{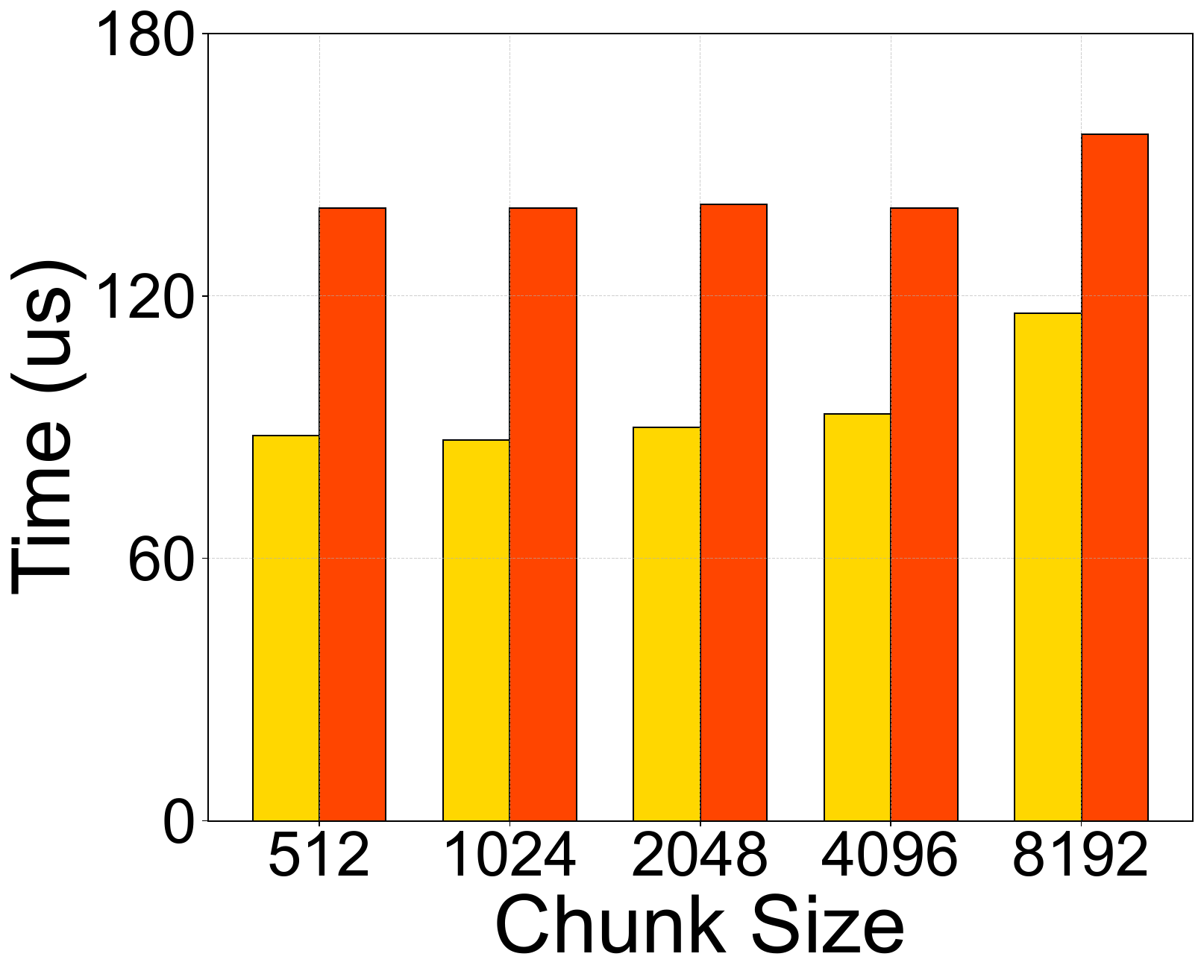}
    \end{subfigure}
    \begin{subfigure}{0.195\linewidth}
        \centering
        \includegraphics[width=\linewidth]{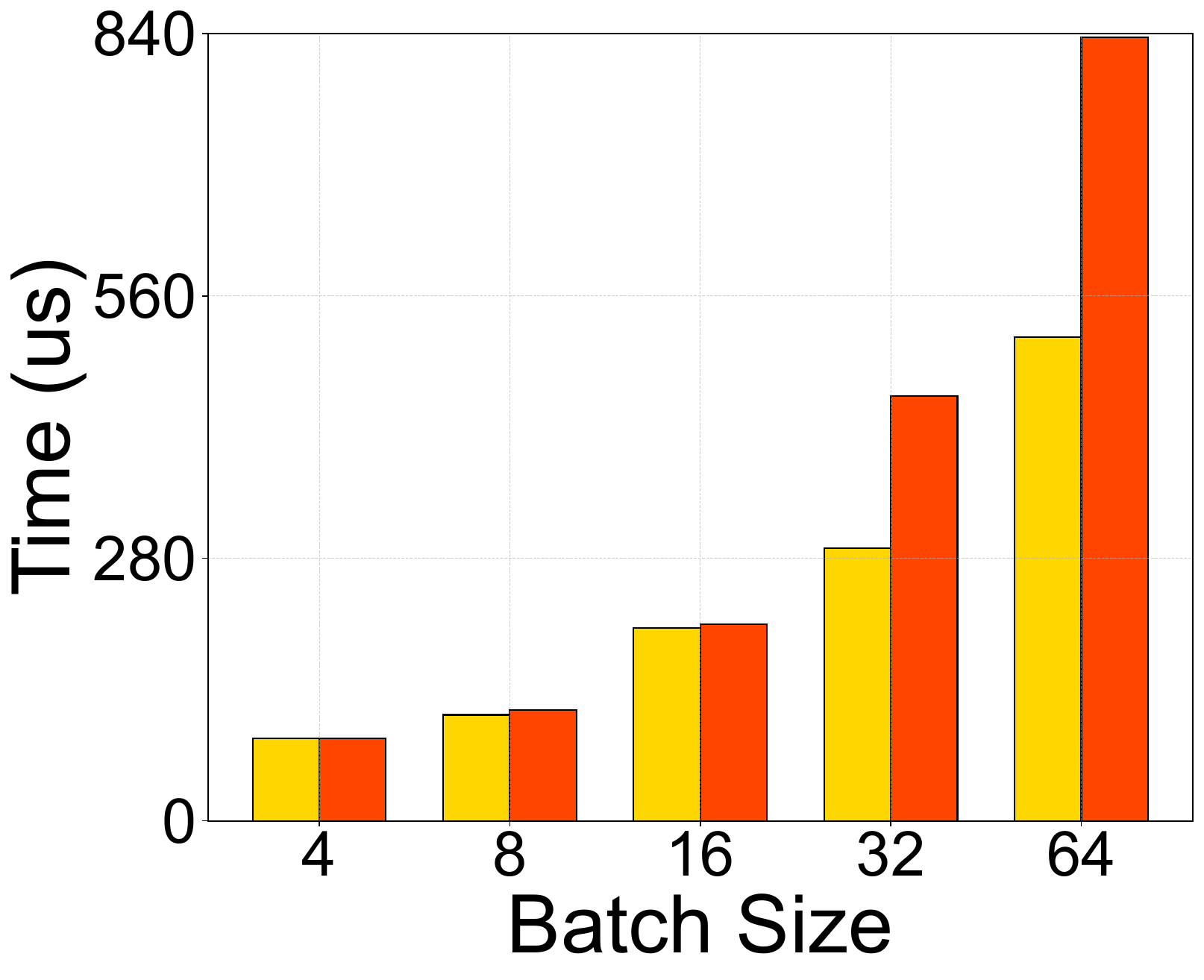}
    \end{subfigure}
    \begin{subfigure}{0.195\linewidth}
        \centering
        \includegraphics[width=\linewidth]{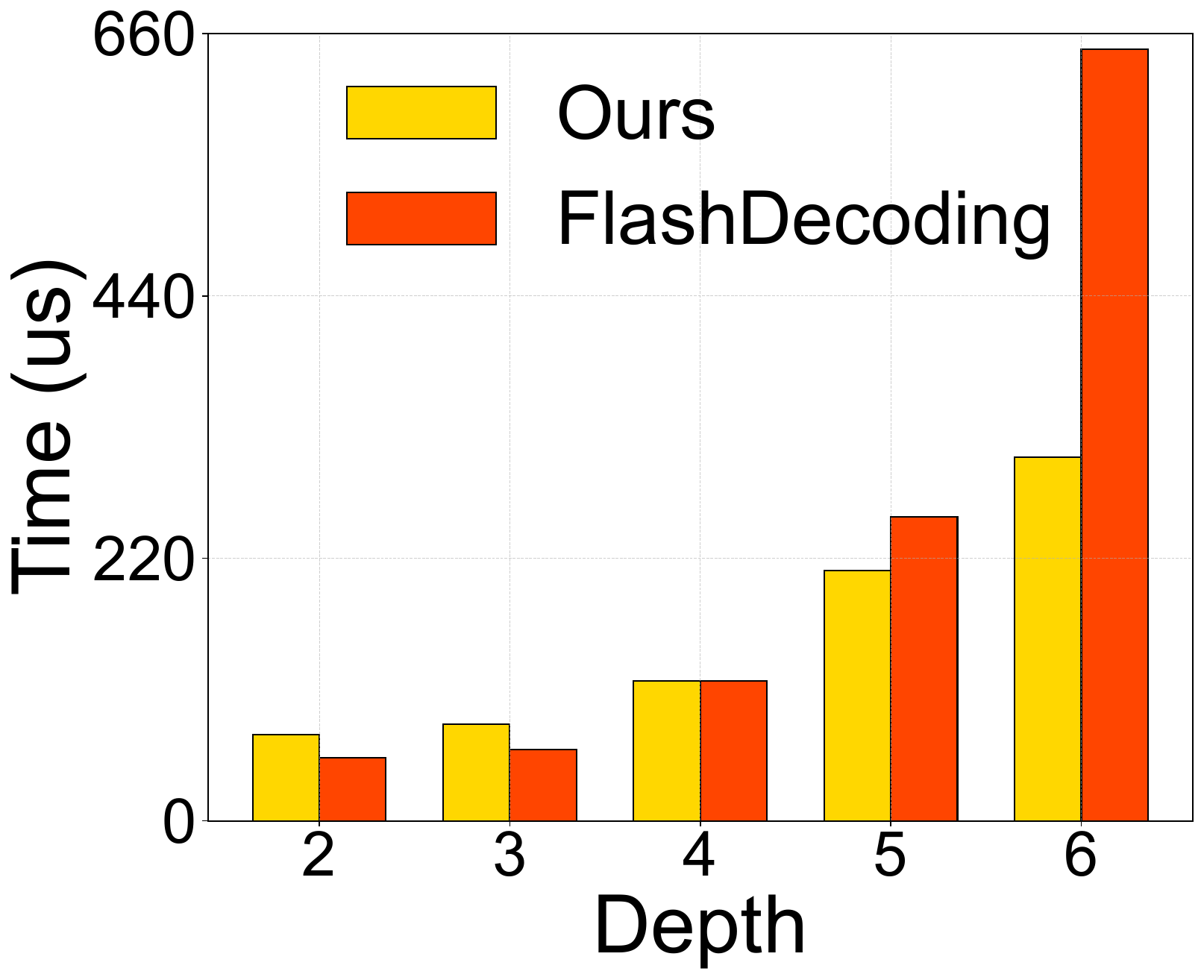}
    \end{subfigure}
    \begin{subfigure}{0.195\linewidth}
        \centering
        \includegraphics[width=\linewidth]{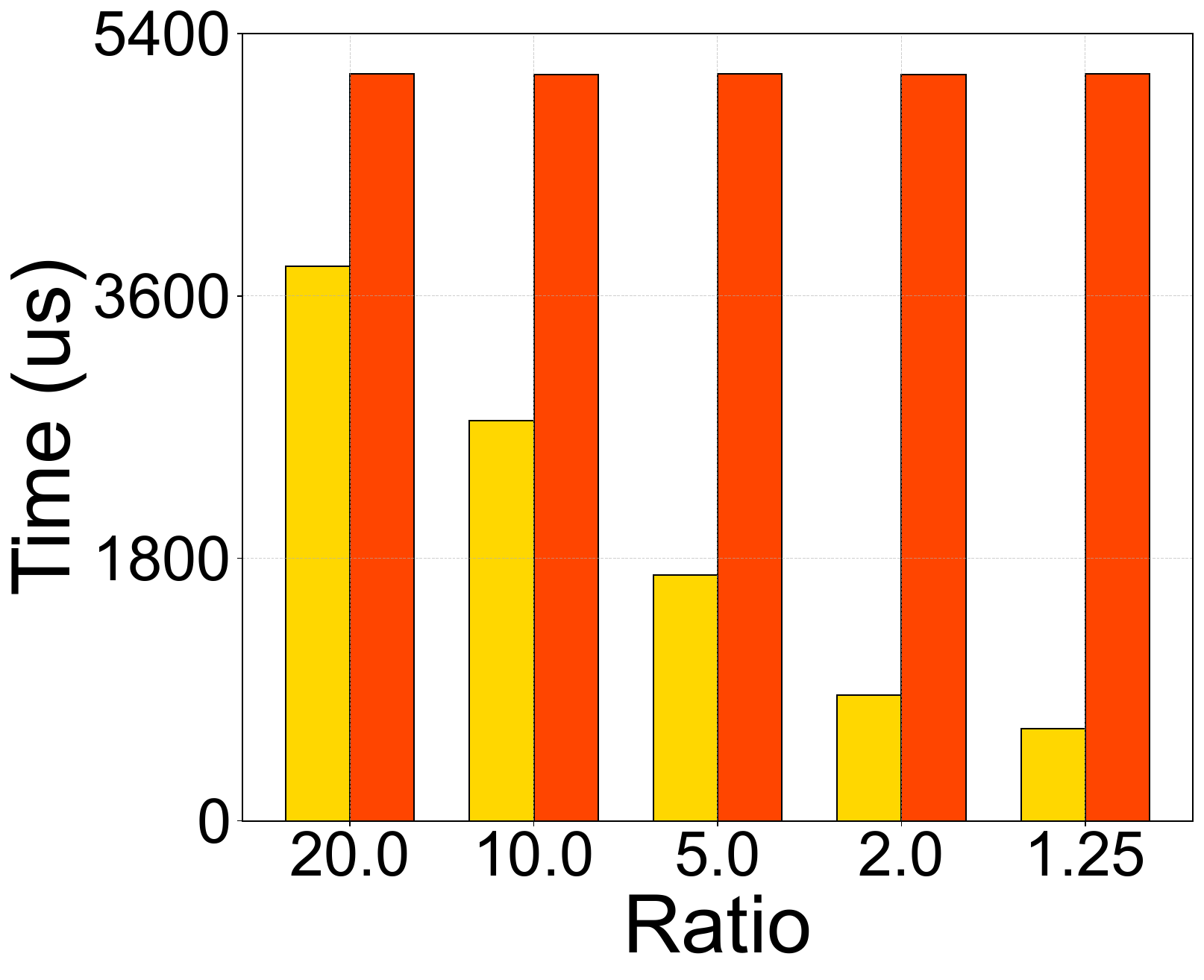}
    \end{subfigure}
    \begin{subfigure}{0.195\linewidth}
        \centering
        \includegraphics[width=\linewidth]{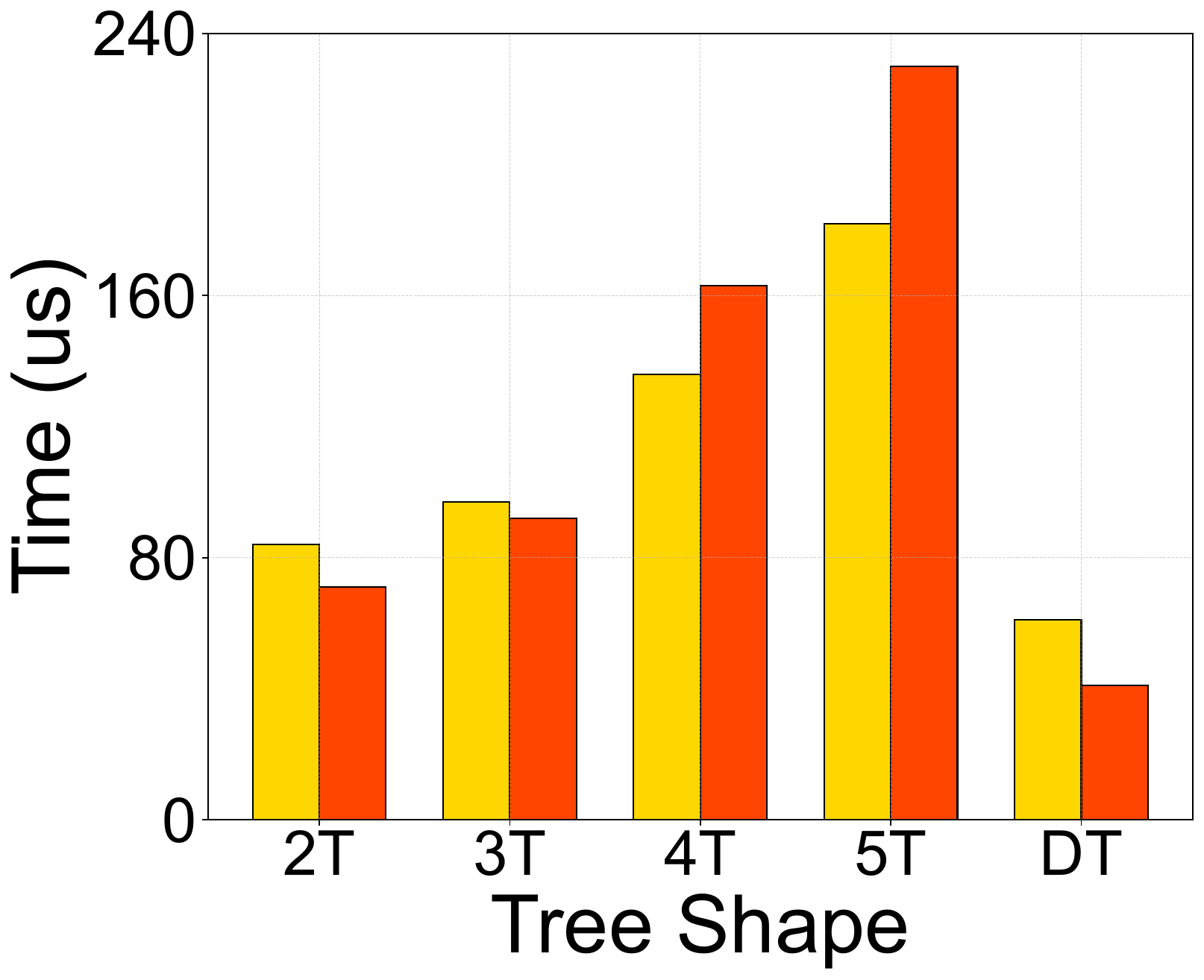}
    \end{subfigure}
    \caption{CoDec vs. FlashDecoding on execution time.}
    \label{fig:comp-sota-attn-time}
\end{figure*}

\begin{figure*}[tbp]
    \centering
    \begin{subfigure}{0.195\linewidth}
        \centering
        \includegraphics[width=\linewidth]{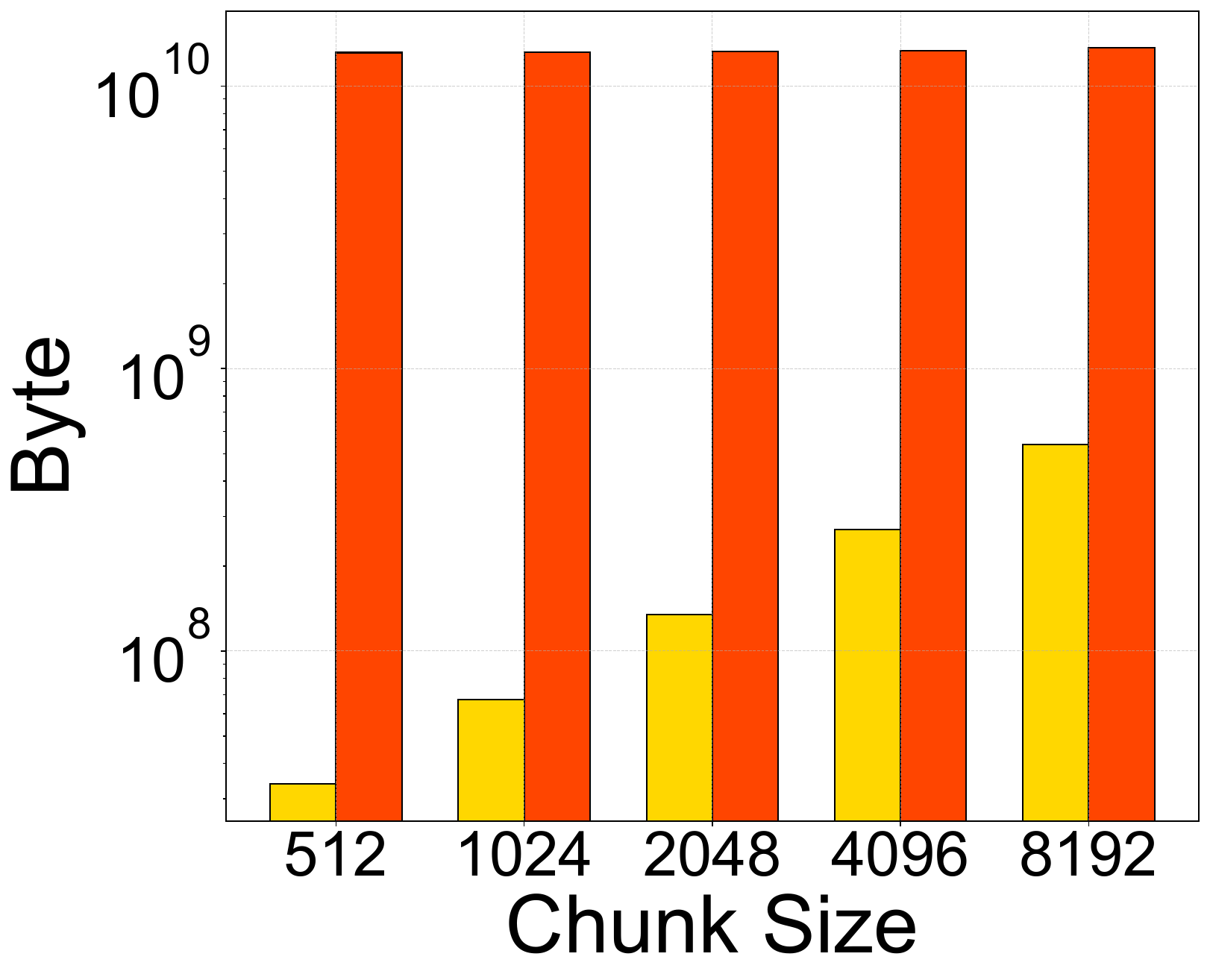}
    \end{subfigure}
    \begin{subfigure}{0.195\linewidth}
        \centering
        \includegraphics[width=\linewidth]{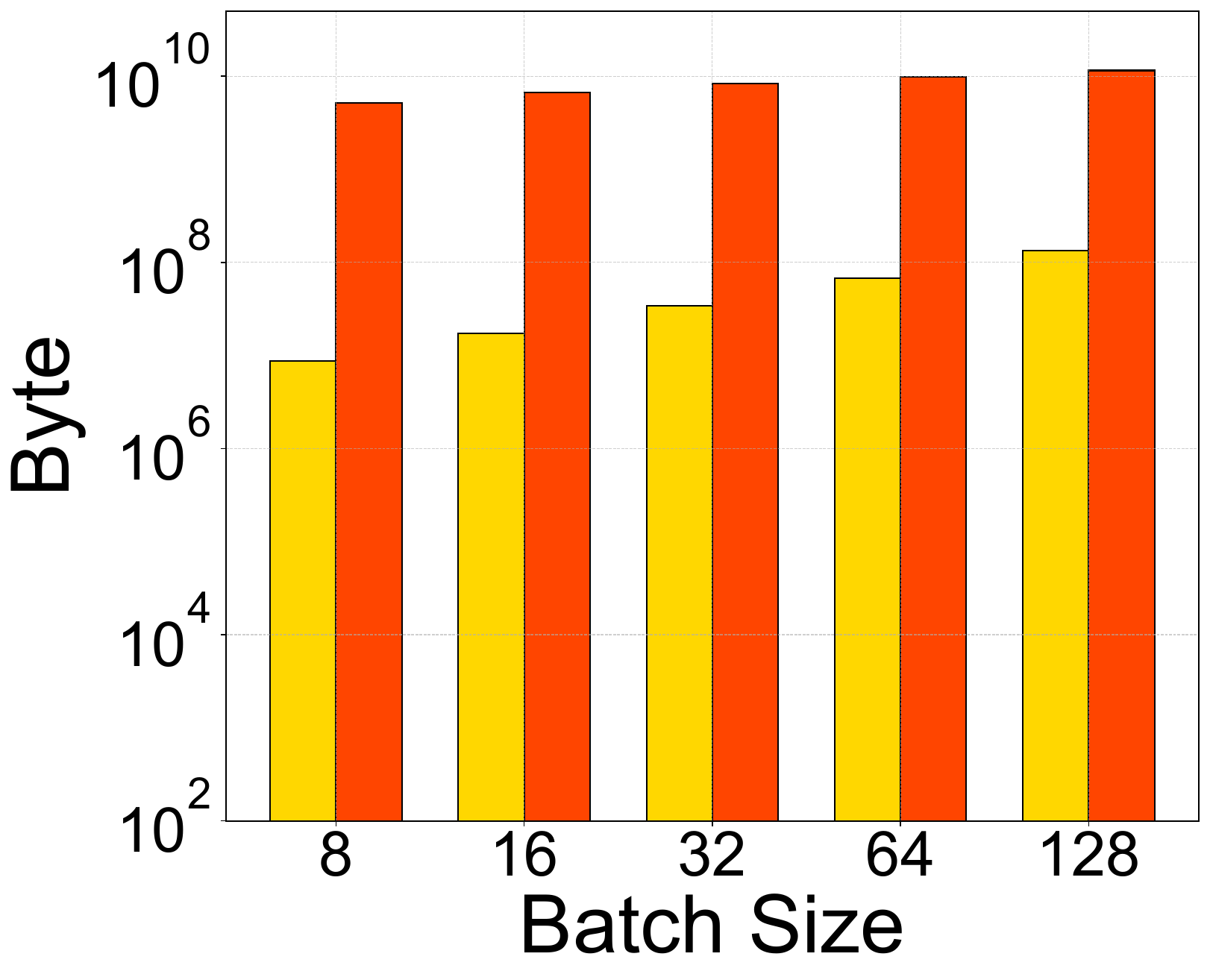}
    \end{subfigure}
    \begin{subfigure}{0.195\linewidth}
        \centering
        \includegraphics[width=\linewidth]{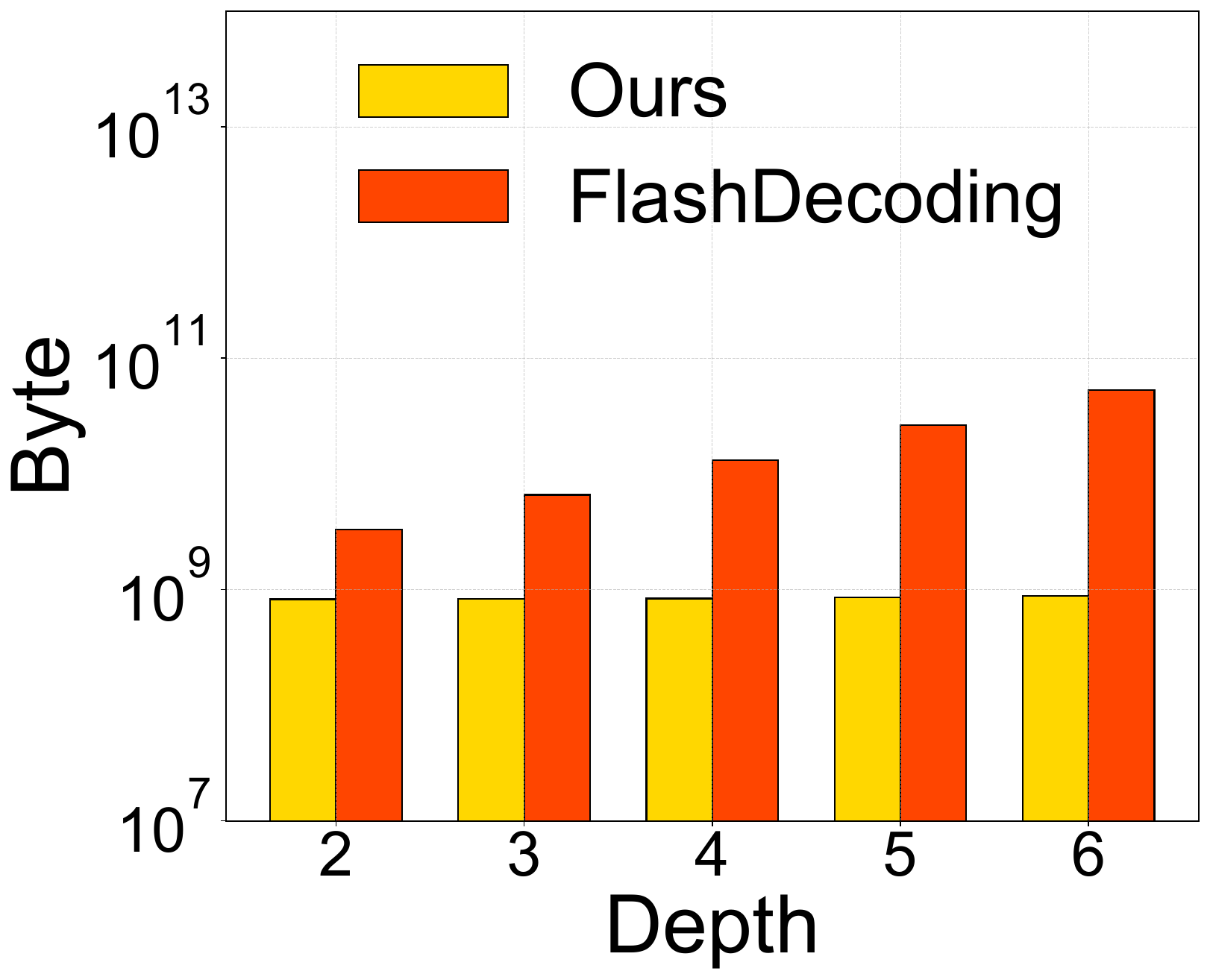}
    \end{subfigure}
    \begin{subfigure}{0.195\linewidth}
        \centering
        \includegraphics[width=\linewidth]{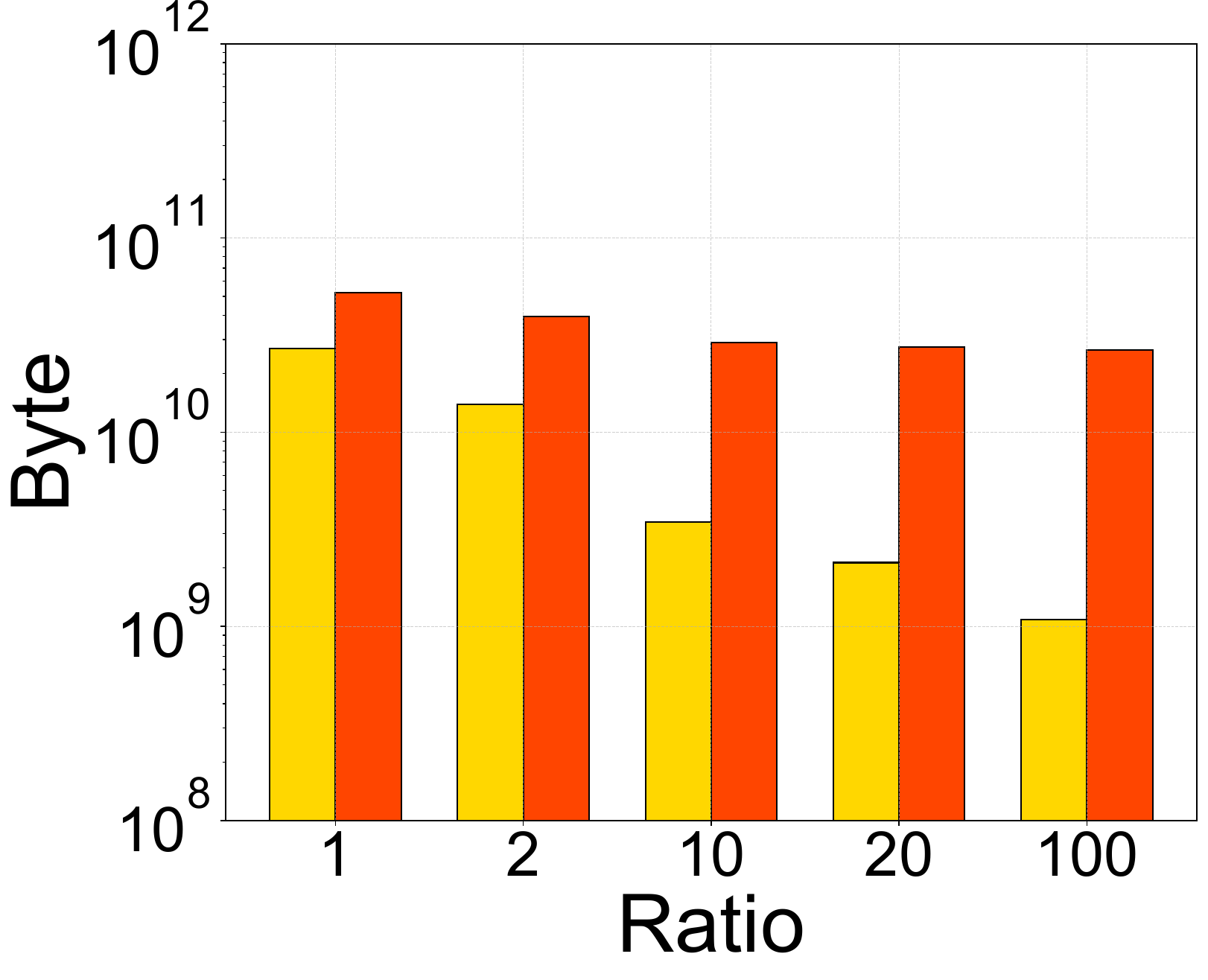}
    \end{subfigure}
    \begin{subfigure}{0.195\linewidth}
        \centering
        \includegraphics[width=\linewidth]{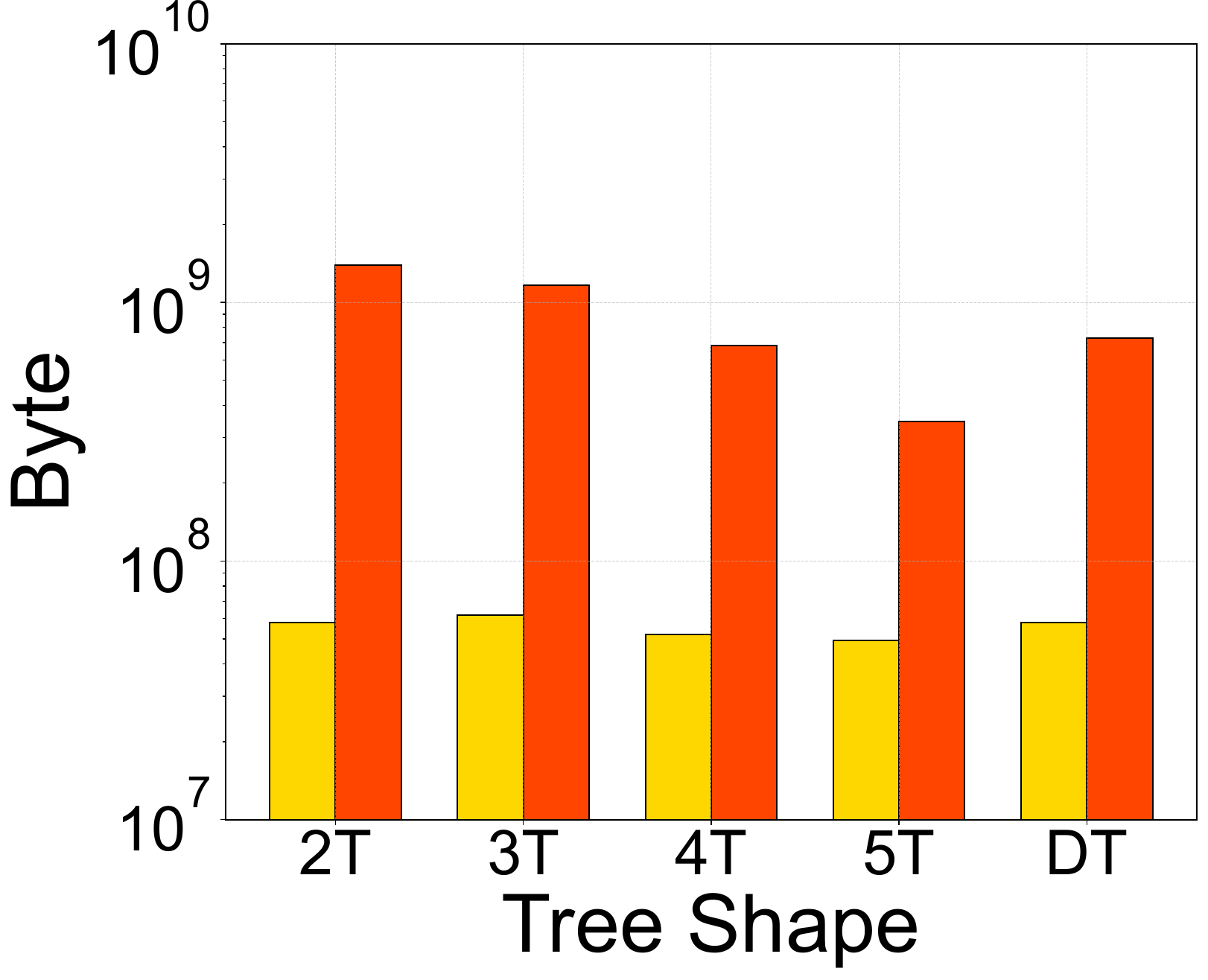}
    \end{subfigure}
    \caption{CoDec vs. FlashDecoding on global memory access.}
    \label{fig:comp-sota-glb-mem}
\end{figure*}

\begin{figure*}[tbp]
    \centering
    \begin{subfigure}{0.195\linewidth}
        \centering
        \includegraphics[width=\linewidth]{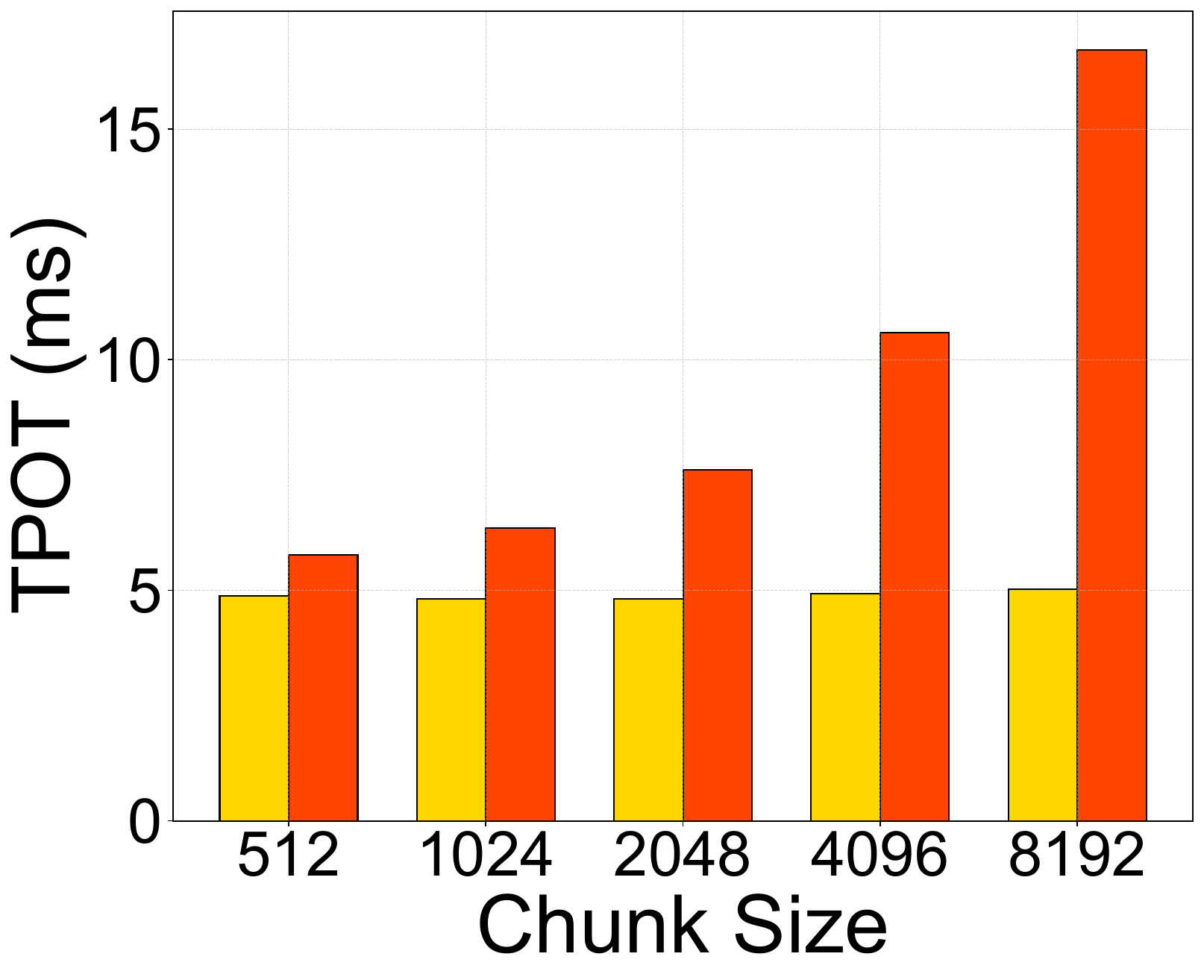}
    \end{subfigure}
    \begin{subfigure}{0.195\linewidth}
        \centering
        \includegraphics[width=\linewidth]{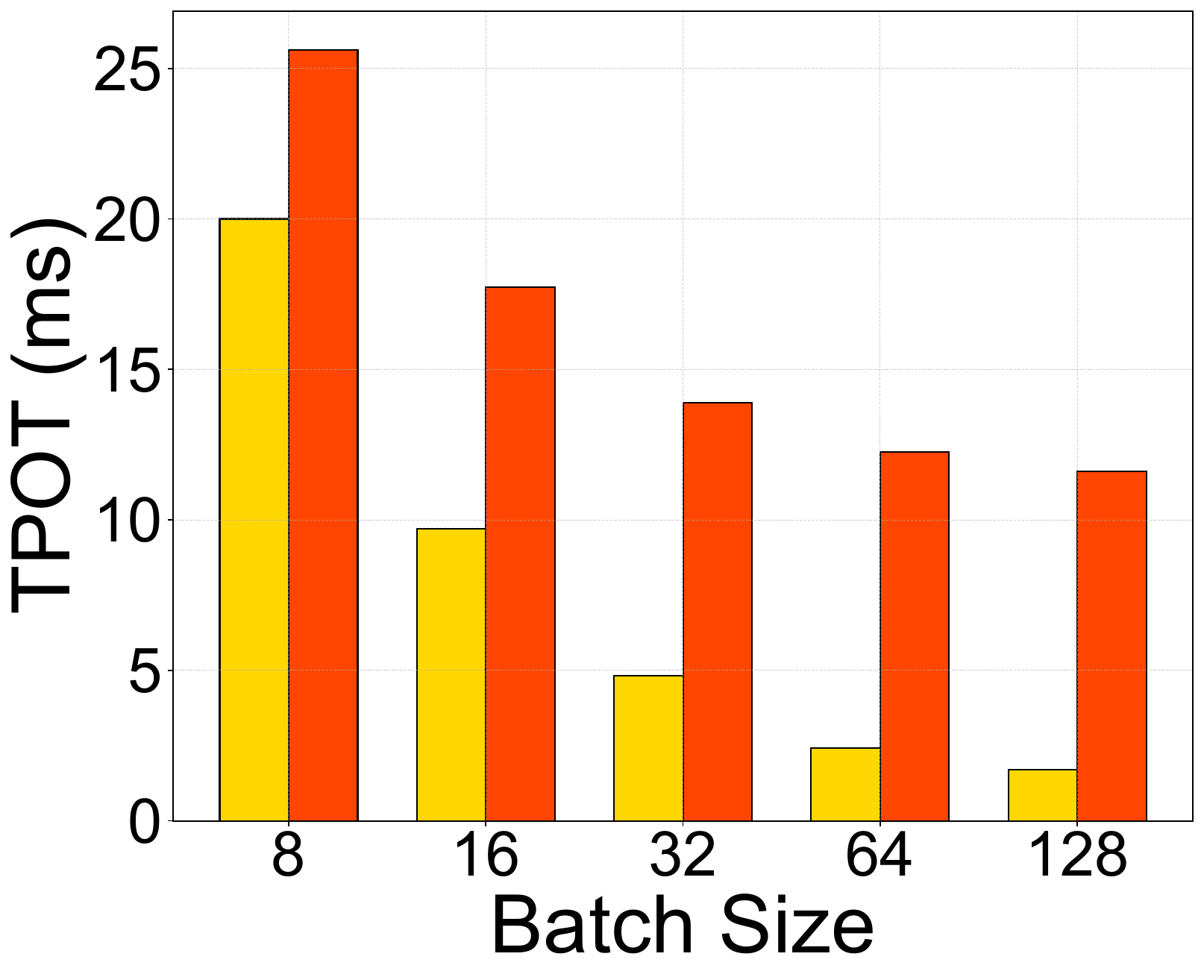}
    \end{subfigure}
    \begin{subfigure}{0.195\linewidth}
        \centering
        \includegraphics[width=\linewidth]{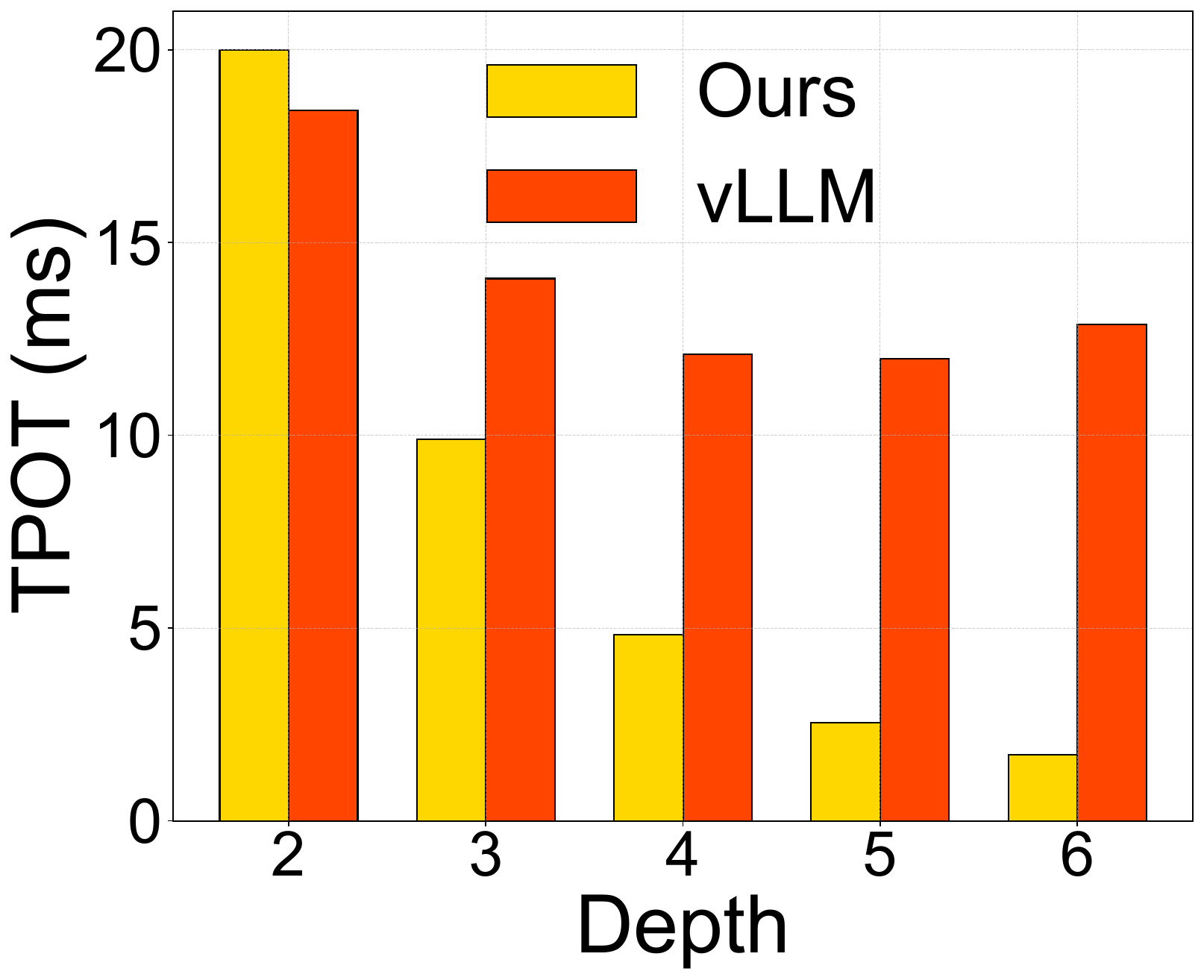}
    \end{subfigure}
    \begin{subfigure}{0.195\linewidth}
        \centering
        \includegraphics[width=\linewidth]{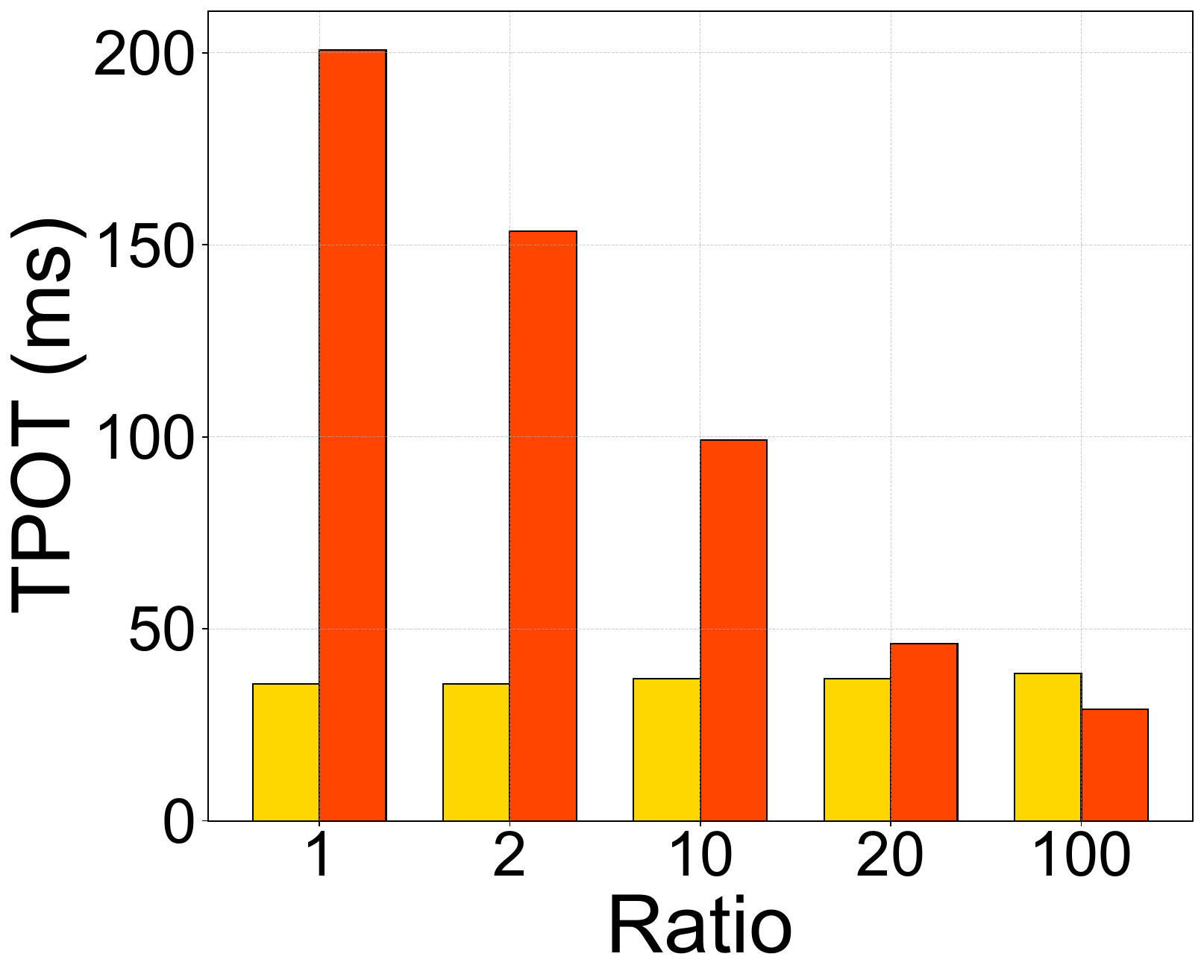}
    \end{subfigure}
    \begin{subfigure}{0.195\linewidth}
        \centering
        \includegraphics[width=\linewidth]{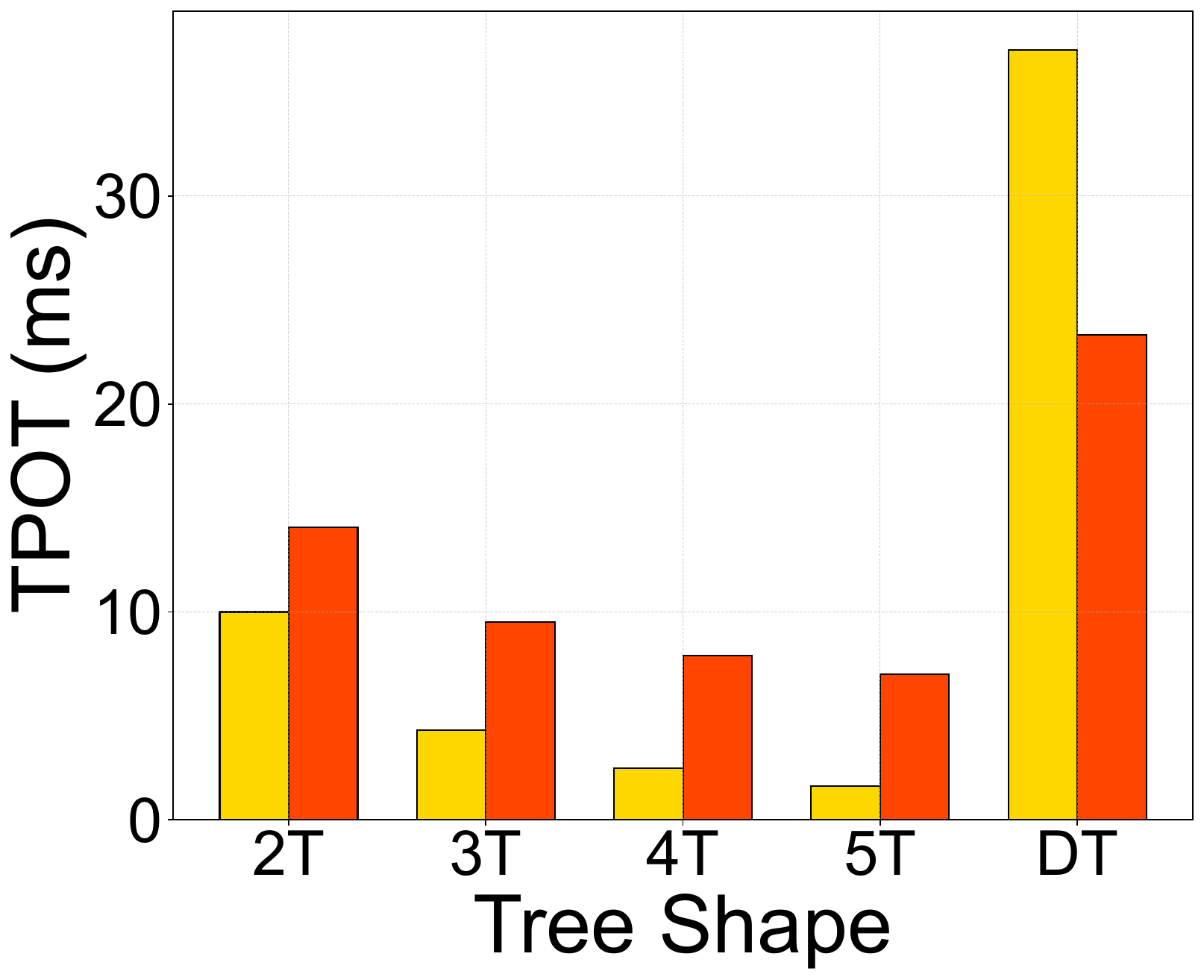}
    \end{subfigure}
    \caption{CoDec vs. vLLM on end-to-end time.}
    \label{fig:comp-sota-e2e-time}
\end{figure*}

\begin{figure}[t]
    \centering

    \begin{subtable}[t]{0.6\columnwidth}
        \vspace{10pt} 
        \centering
        \setlength{\tabcolsep}{7pt} 
        \footnotesize
        \begin{tabular}{@{}c c c c@{}}
            \hline
            \multirow{2}{*}{\textbf{Dataset}} & \multirow{2}{*}{\textbf{Category}} & \textbf{Average} & \multirow{2}{*}{\textbf{Task}} \\
                                              &                                    & \textbf{Tokens}  &                                \\
            \hline
            arXiv                             & \makecell[c]{Phys.,                                                                    \\Math,etc.} & 20,887 & \makecell[l]{Summarization} \\
            \hline
            Wiki                              & \makecell[c]{Hist.,                                                                    \\Polit.,etc.}  & 21,017 & \makecell[l]{short dep. QA \\ long dep. QA} \\
            \hline
            Scripts                           & \makecell[c]{Action,                                                                   \\Drama,etc.}  & 36,412 & \makecell[l]{short dep. Cloze \\ long dep. Cloze} \\
            \hline
        \end{tabular}
        \caption*{(a) LooGLE Dataset}
    \end{subtable}
    \hspace{0.02\columnwidth}
    \begin{subfigure}[t]{0.32\columnwidth}
        \vspace{0pt} 
        \centering
        \includegraphics[width=\linewidth]{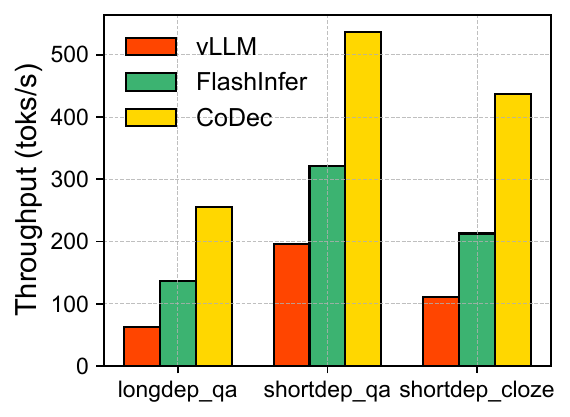}
        \caption*{(b) Throughput}
    \end{subfigure}

    \caption{CoDec vs. SOTA on real-world dataset.}
    \label{fig:dataset-e2e}
\end{figure}

\stitle{Metrics and SOTA.} We evaluate \name in terms of 1) attention-kernel execution time, 2) global memory access in the attention kernel, and 3) end-to-end latency, i.e., TPOT (Time Per Output Token) in decoding.
Regarding attention-kernel execution time and global memory access, we compare \name with FlashDecoding~\cite{dao2023flashattention2} as provided by FlashAttention 2.7.4, which is a strong baseline for long-context decoding. For end-to-end latency, we compare against vLLM 0.6.6~\cite{vllm_sosp23} using a PyTorch 2.6.0 prototype that implements CoDec while following the same paged-KV metadata layout.

\stitle{Attention Execution Time.} As shown in Figure~\ref{fig:comp-sota-attn-time}, \name outperforms FlashDecoding up to 3.6$\times$ and averages 1.9$\times$ speedup across all workloads. We observe that a larger shared prefix results in a more significant speedup, and the case where the shared-to-unique ratio is $100:1$ exhibits the highest speedup. Moreover, given the same shared prefix percentage, the speedup shows a trend of increasing with the decreasing workload size, which is because \name increases the workload in each subtask, resulting in better resource utilization.
Interestingly, irregular workloads, such as 2.9$\times$, exhibit a more pronounced speedup, compared to regular workloads, such as 1.77$\times$. This is because \name can better balance the workload among different subtasks, leading to more efficient resource utilization.

\stitle{Global Memory Access.} Figure~\ref{fig:comp-sota-glb-mem} shows the global memory access of \name and FlashDecoding, which verifies the performance gain of \name. The global memory access of \name is significantly lower than FlashDecoding across all workloads (14.66-409.80$\times$ lower), with an average reduction of 120.85$\times$. Moreover, the same memory reduction does not always lead to the same performance gain, as shown in Figure~\ref{fig:comp-sota-glb-mem} and Figure~\ref{fig:comp-sota-attn-time}, which is attributed to the workload balance and scheduling strategy.

\stitle{End-to-End Latency.}
We also evaluate the end-to-end latency of \name and vLLM in Figure~\ref{fig:comp-sota-e2e-time}.
Our shared prefix contains both coding problems as well as code snippets as few-shot examples of question/answer pairs.
Our benchmark uses the LooGLE dataset to let the model solve competitive programming problems on a single NVIDIA A100 PCIe-40G.
We implement shared prefix tree in PyTorch, caching the shared prefix KV node in each attention layer as its KV cache component.
The end-to-end latency of \name has an average latency reduction of 3.75$\times$ compared to vLLM. We notice that the sequence length has a significant impact on the end-to-end latency. This is due to that the larger the sequence length, the heavier the attention computation, while the FFN computation is insensitive to the sequence length but only related to the batch size.

\rThreeFC{FlashInfer proposes multilevel cascade attention for shared-prefix batch decoding~\cite{cascade-inference}. To compare with this stronger multilevel baseline, we run a micro-benchmark that varies the shared-prefix ratio while keeping the overall context length fixed. Figure~\ref{fig:dataset-e2e} shows that \name consistently achieves lower latency than FlashInfer's multilevel cascade attention across different shared ratios. This advantage mainly comes from two design choices: (1) \name performs workload partitioning with a global view of the entire prefix tree (rather than dividing each prefix node independently), which mitigates imbalance; and (2) \name uses a parallel reduction strategy to aggregate partial results efficiently, avoiding the overhead of launching many reduction kernels when the tree contains a large number of prefix nodes.}

\subsection{Ablation Studies} \label{subsec:ablation}

\begin{figure}[tb]
    \centering
    \vspace{-0.5em}
    \includegraphics[width=\linewidth]{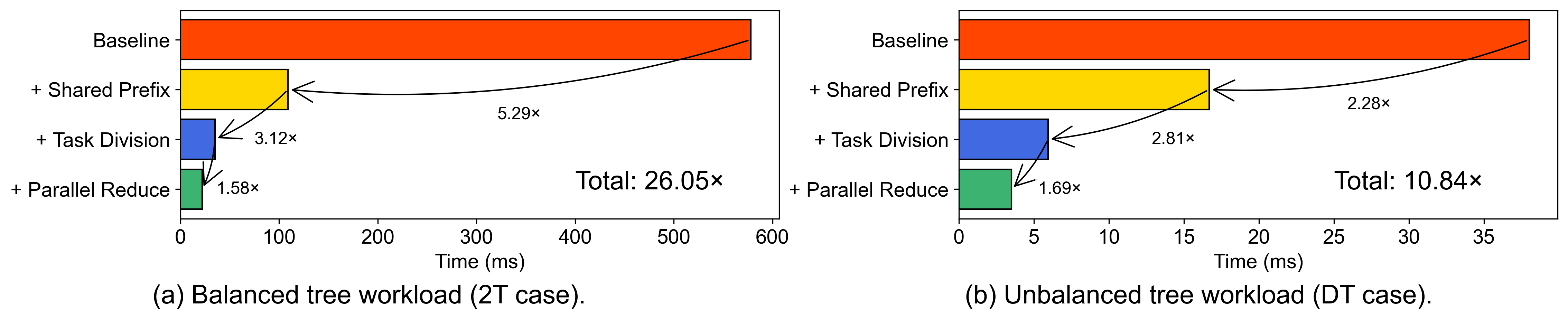}
    \vspace{-1em}
    \caption{Ablation Study.\vspace{-0.5em}}
    \label{fig:ablation}
\end{figure}

To understand how each component contributes to the overall performance gains, we conduct an ablation study that isolates three optimizations:
(1) combining KV-cache accesses via the shared-prefix tree, (2) workload partitioning, and (3) parallel execution and reduction.
We evaluate on two representative workloads: a balanced full binary shared-prefix tree and an unbalanced degenerate tree, both with a maximum context length of 200k tokens.

The results are shown in Figure~\ref{fig:ablation}(a) and Figure~\ref{fig:ablation}(b).
For the unbalanced tree workload, the latency drops from 38.0\,ms without optimization to 3.5\,ms with all optimizations applied, achieving a 10.8$\times$ speedup. Using only the prefix tree or partitioning yields 16.7\,ms and 5.9\,ms, respectively.
For the balanced tree workload, the latency is reduced from 578\,ms to 22.2\,ms with all optimizations, resulting in a 26.1$\times$ speedup. Applying only the prefix tree or partitioning leads to 109.2\,ms and 34.9\,ms, respectively.

These results show that each technique contributes to reducing latency, and combining all three yields the largest speedup.
Notably, the impact of workload balancing and parallelism is more significant for the balanced tree, due to its higher intrinsic computational load.

\subsection{Impact of Division Granularity}
\label{subsec:division-granularity}

To further reveal the impact of task-division granularity, we compare \name against a naive strategy that evenly splits each task into a fixed number of subtasks.
This naive strategy ignores workload skew in the KV-cache tree and query distribution, and thus may either create imbalance (too few splits) or introduce excessive overhead (too many splits).

\begin{figure}[htb]
    \centering
    \includegraphics[width=0.6\linewidth]{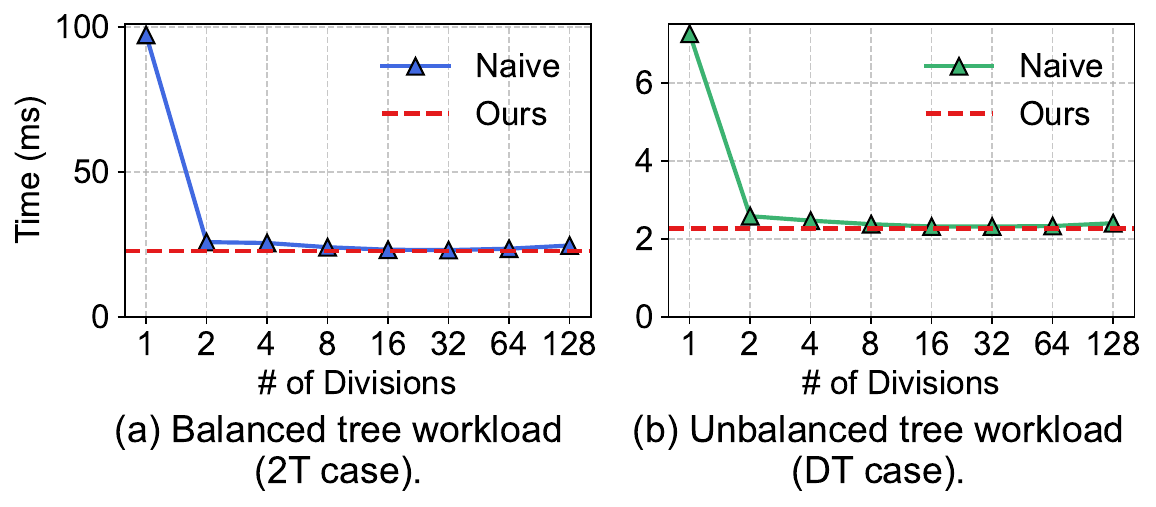}
    \vspace{-1em}
    \caption{Impact of division granularity.}
    \label{fig:split-scaling}
\end{figure}

Figure~\ref{fig:split-scaling} compares the naive approach under different division counts with \name's adaptive division and scheduling.
We plot \name as a horizontal dashed line since it automatically determines an effective division granularity based on the observed workload distribution.
When the division count is 1, the naive approach degenerates to the baseline that executes without any task division.

The experimental results show that our approach outperforms the best fixed-division strategy of the naive approach by 1.02-1.04$\times$ across the tested workloads. Compared to the naive approach without division, our approach achieves 3.20-4.39$\times$ speedup and averages 3.80$\times$ across all workloads. This demonstrates the effectiveness of our division and scheduling strategy in balancing the workload among different subtasks and reducing the overhead of kernel launch and synchronization.

\subsection{Overhead of Task Division}
\label{subsec:task-division-overhead}

\begin{figure}[htb]
    \centering
    \includegraphics[width=0.5\linewidth]{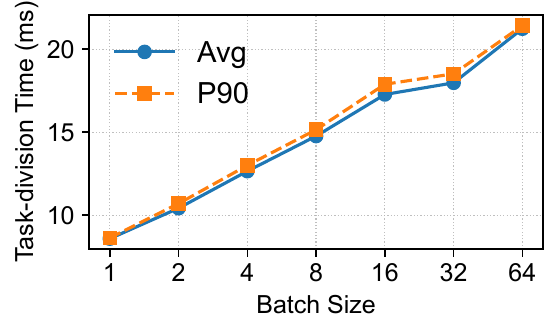}
    \vspace{-1em}
    \caption{CPU overhead of computing the task-division plan as batch size increases.}
    \label{fig:task-division-overhead}
\end{figure}

\rAllFC{Task division is computed on CPU and could become a latency concern when the shared-prefix ratio is low and the attention kernel is relatively fast.
    We therefore measure the end-to-end time of generating a division plan (i.e., cost estimation, pruning, and greedy scheduling) as we increase batch size.
    Figure~\ref{fig:task-division-overhead} shows that the overhead grows with batch size due to the increased number of tasks induced by larger prefix trees, but remains within tens of milliseconds even at batch size 64.
    In practice, we further amortize this cost by reusing a division plan for multiple decoding steps (Section~\ref{sec:implementation}), keeping the partitioning overhead small relative to attention execution time.}

\subsection{Performance in Varying GPU Cards}
\label{subsec:gpu-varing}

To evaluate cross-platform efficiency and performance consistency, we conduct experiments
with a context length of 50K tokens on five modern GPUs.

\begin{figure}[t]
    \centering
    \includegraphics[width=.7\columnwidth]{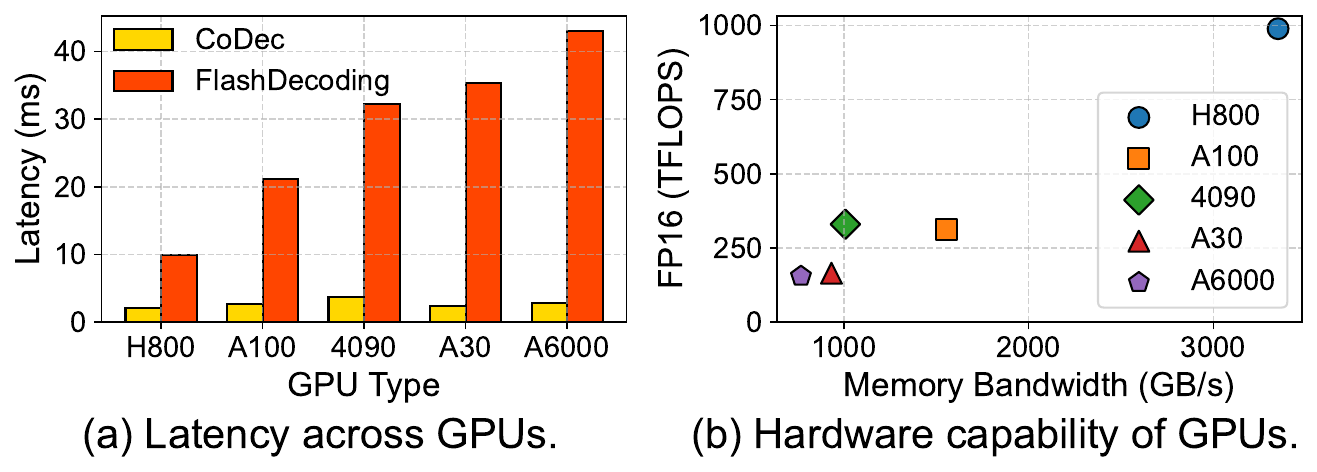}
    \vspace{-1em}
    \caption{Performance on diverse GPUs.}
    \label{fig:gpu_varing}
\end{figure}

As shown in Figure~\ref{fig:gpu_varing}, \name consistently outperforms FlashDecoding~\cite{flashdecoding} across all tested GPUs. On the H800, our method achieves a latency of 2.094\,ms, compared to 9.900\,ms for FlashDecoding, resulting in a 4.7$\times$ speedup. Even on lower-end GPUs such as the A6000, \name maintains a 15$\times$ advantage (2.869\,ms vs. 43.048\,ms).

The performance gap notably widens on GPUs with lower memory bandwidth.
For example, FlashDecoding suffers on the A6000 (768\,GB/s bandwidth), whereas our method degrades much more gracefully.
This indicates that \name is less sensitive to hardware limitations, making it suitable for deployment across both data-center and consumer-grade GPUs.

\subsection{Performance in Varying Models}
\label{subsec:attention-variants}

\stitle{Attention Variants.} \rAllFC{CoDec is designed to be compatible with widely used attention variants in modern LLMs, including MHA, MQA, and GQA. To evaluate this, we vary the grouping configuration of GQA (i.e., how many query heads share one KV head) and compare against FlashDecoding under the same shared-prefix workload.}

\begin{figure}[htbp]
    \centering
    \begin{subfigure}[t]{0.35\columnwidth}
        \centering
        \includegraphics[width=\linewidth]{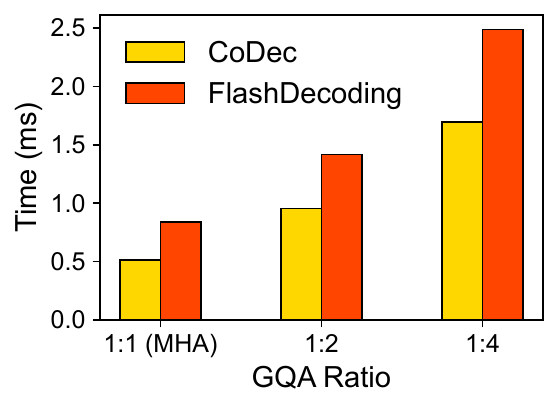}
        \subcaption{Attention variants.}
        \label{fig:attention_variants}
    \end{subfigure}
    \begin{subfigure}[t]{0.35\columnwidth}
        \centering
        \includegraphics[width=\linewidth]{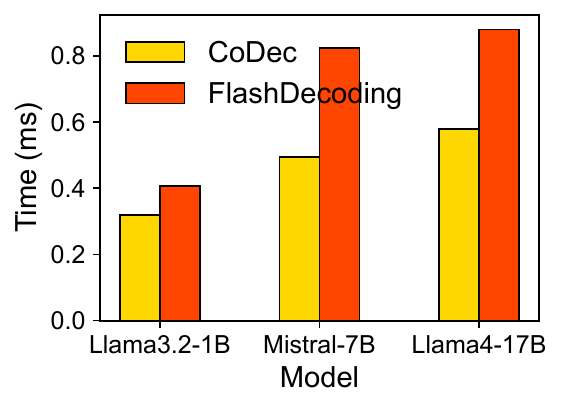}
        \subcaption{Different models.}
        \label{fig:model_varying}
    \end{subfigure}
    \vspace{-0.8em}
    \caption{Performance under different attention variants and models.
        \vspace{-0.5em}}
    \label{fig:attn_model_varying}
\end{figure}

\rAllFC{As shown in Figure~\ref{fig:attention_variants}, CoDec consistently reduces decoding latency across the tested configurations. The gain remains stable as the group size changes, indicating that our kernel design is not tied to a specific head layout and can generalize to different KV-sharing patterns encountered in real models.}

\stitle{Model Sizes.} \rAllFC{We further test CoDec on multiple representative model families/sizes with different head configurations. Figure~\ref{fig:model_varying} shows that CoDec maintains consistent latency reduction across models, suggesting that the benefit mainly comes from eliminating redundant KV-cache reads and improving workload scheduling, rather than relying on a particular architecture.}

\section{Related Work}
\label{sec:related_work}

\stitle{Prefix caching.}
Prefix sharing is widely used to avoid redundant prefill computation by reusing KV states for common prompts.
PagedAttention~\cite{vllm_sosp23} provides a paged KV-cache abstraction that enables efficient memory management and sharing across requests.
vLLM further introduces automatic prefix caching (APC)~\cite{vllm_apc} to retain hot prefixes, while SGLang~\cite{zheng2024sglangefficientexecutionstructured} uses a radix-tree-like structure to reuse historical KV states across structured program executions.
\rAllFC{Recent data-management work also studies KV-cache reuse policies and chunk-level caching for LLM serving.
HotPrefix~\cite{hotprefix} schedules which prefix KV states to keep and when to promote them across GPU/CPU memory, while Cache-Craft~\cite{cachecraft} manages reusable chunk-caches for retrieval-augmented generation (RAG) workloads.
These systems focus on improving KV reuse and cache management primarily in the prefill phase; \name is complementary and targets the decode phase, where attention becomes memory-bound and shared-prefix KV reads can be combined.}

\stitle{Prefix-tree decoding and multilevel attention.}
Several recent works explore decoding-time attention on prefix trees.
ChunkAttention~\cite{ye2024chunkattentionefficientselfattentionprefixaware} and DeFT~\cite{yao2025deftdecodingflashtreeattention} consider tree-structured inference and propose prefix-aware attention execution strategies.
\rThreeFC{FlashInfer proposes multilevel cascade attention for shared-prefix batch decoding~\cite{cascade-inference}.
\name differs from these approaches in two key aspects: (1) it performs workload partitioning with a global view of the entire prefix tree to mitigate imbalance, rather than treating each prefix node independently; and (2) it uses a parallel, tree-structured reduction strategy to aggregate partial results efficiently, avoiding the overhead of launching many reduction kernels when the tree contains a large number of nodes.}

\stitle{Other attention mechanisms.}
In addition to multi-head attention (MHA), 
Multi-query attention (MQA)~\cite{shazeer2019fasttransformerdecodingwritehead}, Grouped-query attention (GQA)~\cite{ainslie2023gqatraininggeneralizedmultiquery},
other advanced attention mechanisms have been proposed.
\cite{deepseekai2024deepseekv2strongeconomicalefficient, deepseekai2025deepseekv3technicalreport} propose multi-head latent attention (MLA), a novel paradigm that projects queries, keys, and values into a low-dimensional latent space across multiple heads.
\rAllFC{\name can support MHA/MQA/GQA directly and can be extended to MLA by reconstructing per-head KV blocks before applying the same prefix-aware attention and reduction pipeline.}

\stitle{Distributed KV cache management.}
PagedAttention~\cite{vllm_sosp23} integrates the paged memory management mechanism into attention computation, mitigating memory fragmentation and enhancing inference throughput. \cite{qin2024mooncakekvcachecentricdisaggregatedarchitecture,hu2024memservecontextcachingdisaggregated} explore a distributed environment, employing a memory pool for KV caching across multiple instances. Specifically, \cite{qin2024mooncakekvcachecentricdisaggregatedarchitecture} utilizes hashing, while \cite{hu2024memservecontextcachingdisaggregated} adopts a global prompt tree, to retrieve historical KV cache. We notice that \name is orthogonal to these approaches, as it considers the KV cache of runtime attention computation on a single instance, and can be combined with these storage-level optimizations.

\stitle{Distributed attention computation.}
With the rapid growth of model sizes and sequence lengths, the need for distributed attention computation has become increasingly important.
When employing traditional parallelization methods to distribute attention computation,
tensor parallelism~\cite{shoeybi2020megatronlmtrainingmultibillionparameter} is employed on the head dimension, while data parallelism~\cite{pope2022efficientlyscalingtransformerinference} partitions the batch dimension.
Recently, for scenarios involving long sequence lengths, sequence parallelism~\cite{yang2024contextparallelismscalablemilliontoken,wu2024loongserveefficientlyservinglongcontext,li2022sequenceparallelismlongsequence}, involves partitioning along the sequence dimension.
\name can be easily integrated with tensor parallelism as head dimension do not affect our design, while the sequence parallelism and data parallelism may lead to a lower sharing ratio, which is an interesting direction to explore the task division in these distributed settings.

\section{Conclusion}

In this paper, we presented CoDec, a dedicated prefix-shared decoding operator designed to significantly accelerate attention computation, which dominates the memory-bound LLM decode stage as the primary performance bottleneck, by efficiently leveraging shared KV cache patterns across multiple requests.
Our approach introduces two key innovations: (1) a novel shared-prefix attention kernel that optimizes memory hierarchy through sophisticated indexing between prefix KV cache trees and query tensors while exploiting both intra-block and inter-block parallelism; and (2) a comprehensive workload balancing mechanism featuring a profile-based cost estimator, intelligent task division, and efficient scheduling algorithms to handle irregular workloads.
Experimental results show that CoDec achieves significant performance improvements over state-of-the-art FlashDecoding kernels, with up to 11.56$\times$ speedup and 150.56$\times$ memory access reduction across diverse workloads.

\begin{acks}
  This work was supported by the Key Program of the National Natural Science Foundation of China under Grant Nos. 62325205, 62502193 and 62272215, the Natural Science Foundation of Jiangsu Province under Grant Nos. BK20243053, Fundamental and Interdisciplinary Disciplines Breakthrough Plan of the Ministry of Education of China (No. JYB2025XDXM901), the Nanjing ``U35'' Talent Cultivation Program (No. U (2024) 001), Nanjing Kunpeng\&Ascend Center of Cultivation, and Nanjing University-China Mobile Communications Group Co., Ltd. Joint Institute. Rong Gu is the corresponding author.
\end{acks}


\bibliographystyle{ACM-Reference-Format}
\bibliography{ref.bib}










\end{document}